\documentclass[11pt]{article}

\usepackage[preprint]{acl}

\usepackage{times}
\usepackage{latexsym}

\usepackage{booktabs}
\usepackage{hyperref}
\usepackage{url}
\usepackage{graphicx}
\usepackage{subcaption}
\usepackage{booktabs}
\usepackage{amsmath}
\usepackage{cleveref}
\usepackage{graphicx}
\usepackage{verbatim}
\usepackage{wrapfig}
\usepackage{lipsum}
\usepackage{xcolor}  
\usepackage{diagbox}
\usepackage{multirow}
\usepackage{multicol}
\usepackage{amssymb}% http://ctan.org/pkg/amssymb
\usepackage{pifont}% http://ctan.org/pkg/pifont
\usepackage{amsfonts}
\newcommand{\cmark}{\ding{51}}%
\newcommand{\xmark}{\ding{55}}%
\usepackage{tcolorbox}

\usepackage[T1]{fontenc}
\usepackage[utf8]{inputenc}

\usepackage{microtype}

\usepackage{inconsolata}
\usepackage{placeins}   % for \FloatBarrier
\usepackage{float}      % for [H], only when you really want exact placement
\usepackage{graphicx}

\tcbset{
  promptbox/.style={
    sharp corners, boxsep=0pt, boxrule=0.6pt,
    colback=white, colframe=black!70,
    colbacktitle=black!70, coltitle=white, titlerule=0pt,
    fonttitle=\footnotesize\bfseries,
    left=5pt, right=5pt, top=4pt, bottom=4pt,
    toptitle=2.5pt, bottomtitle=2.5pt,
  }
}
\usepackage{arydshln}
\definecolor{e7}{RGB}{104, 193, 109}
\definecolor{e8}{RGB}{117, 198, 115}
\definecolor{e9}{RGB}{134, 203, 126}
\definecolor{e10}{RGB}{151, 208, 137}
\definecolor{e11}{RGB}{166, 213, 148}
\definecolor{e12}{RGB}{180, 218, 159}
\definecolor{e13}{RGB}{194, 223, 171}
\definecolor{e14}{RGB}{206, 227, 182}
\definecolor{e15}{RGB}{217, 232, 194}
\definecolor{e16}{RGB}{227, 237, 205}
\definecolor{e17}{RGB}{236, 241, 217}
\definecolor{e18}{RGB}{243, 246, 229}
\definecolor{e19}{RGB}{249, 250, 241}
\definecolor{e20}{RGB}{254, 254, 253}

\definecolor{r7}{RGB}{132,159,219}
\definecolor{r8}{RGB}{140,164,223}
\definecolor{r9}{RGB}{148,169,226}
\definecolor{r10}{RGB}{157,174,230}
\definecolor{r11}{RGB}{166,180,233}
\definecolor{r12}{RGB}{175,187,236}
\definecolor{r13}{RGB}{184,193,239}
\definecolor{r14}{RGB}{193,200,242}
\definecolor{r15}{RGB}{202,208,244}
\definecolor{r16}{RGB}{212,216,247}
\definecolor{r17}{RGB}{222,224, 249}
\definecolor{r18}{RGB}{232,233,251}
\definecolor{r19}{RGB}{242,242, 253}
\definecolor{r20}{RGB}{252, 252, 255}

\title{Do Language Models Reason Across Languages?}

\author{
    Yan Meng \quad Wafaa Mohammed \quad Christof Monz \\
    Language Technology Lab\\
    University of Amsterdam\\
    Science Park \texttt{904}, The Netherlands \\
    \texttt{\{y.meng\}@uva.nl}
}

\begin{document}

\maketitle
\begin{abstract}

Real-world data sources are inherently multilingual, which raises the question \textit{whether} language models can synthesize information across languages. 
In this paper, we introduce a simple two-hop question answering setting, where answering a question requires making inferences over two documents in different languages. 
We find that language models are more sensitive to language variation in answer-span documents than in those providing bridging information, despite the equal importance of both documents for answering a question. 
Under a step-by-step sub-question evaluation, we further show that in up to 33\% of multilingual cases, models fail to infer the bridging information in the first step yet still answer the overall question correctly.
This indicates that reasoning in language models, especially in multilingual settings, does not follow a faithful step-by-step decomposition.
%Subsequently, we show that the absence of reasoning decomposition leads to around $18\%$ composition failure, where both sub-questions are answered correctly but fail for the final two-hop questions. 
%To mitigate this, we propose a simple three-stage \textsc{SubQ} prompting method to guide the multi-step reasoning with sub-questions, which boosts accuracy from $10.1\%$ to $66.5\%$.
Moreover, language models also struggle to compose intermediate results: in up to 18\% of cases, they answer both sub-questions correctly but fail to derive the correct final answer. 
To examine whether explicit decomposition can help, we evaluate a three-stage \textsc{SubQ} prompting setup that provides the relevant intermediate sub-questions during inference. 
This improves two-hop QA performance on both the compositional failure cases and the full evaluation set.

\end{abstract}

\section{Introduction}

Reasoning is a central aspect of human cognition and refers to the process of drawing new conclusions by combining multiple pieces of evidence through logical inference \citep{KURTZ1999145}. 
Large language models (LLMs) have demonstrated strong reasoning performance across pieces of evidence within the same language, which almost exclusively is English  \citep{liu2025logicalreasoninglargelanguage}. 
However, real-world information is inherently multilingual and distributed across languages. 
Enhancing multilingual reasoning is therefore essential for building globally reliable AI systems.
This requires models not only to comprehend information in multiple languages but also to integrate knowledge across them \citep{Ghosh2025TheMM}.

Recent works \citep{qin-etal-2023-cross, huang-etal-2023-languages, xcot} explore multilingual reasoning in non-linguistic domains such as mathematics and coding, which inherently require multi-step reasoning. 
However, their evaluation is solely based on final performance, e.g., accuracy, which cannot truly capture whether language models perform inference across multiple multilingual sources or instead rely on shortcut heuristics.
 In this work, we focus on a \textit{decomposed} two-hop question answering task that requires chaining information in two contexts to answer a question. 
The controlled decomposition of this task enables fine-grained evaluation of multilingual reasoning at each step.

%While extensively studied in monolingual English settings \citep{yang-etal-2024-large-language-models,biran-etal-2024-hopping,yu-etal-2025-unleashing}, it remains unclear how well LLMs perform such reasoning across multilingual documents -- an important capability for real-world applications where information is naturally distributed across different cultural sources.
%However, far less is understood about multilingual multi-step reasoning in linguistic domains, which requires analyzing, synthesizing, and drawing inferences solely from textual information. 
%Multi-hop reasoning is a representative task that requires integrating information across multiple contexts to answer a complex question. 
%While extensively studied in monolingual English settings \citep{yang-etal-2024-large-language-models,biran-etal-2024-hopping,yu-etal-2025-unleashing}, it remains unclear how well LLMs perform multi-hop reasoning across multilingual documents -- an important capability for real-world applications where information is naturally distributed across different languages and cultural sources.

\begin{figure*}[!t]
    \centering
        \includegraphics[width=0.95\textwidth]{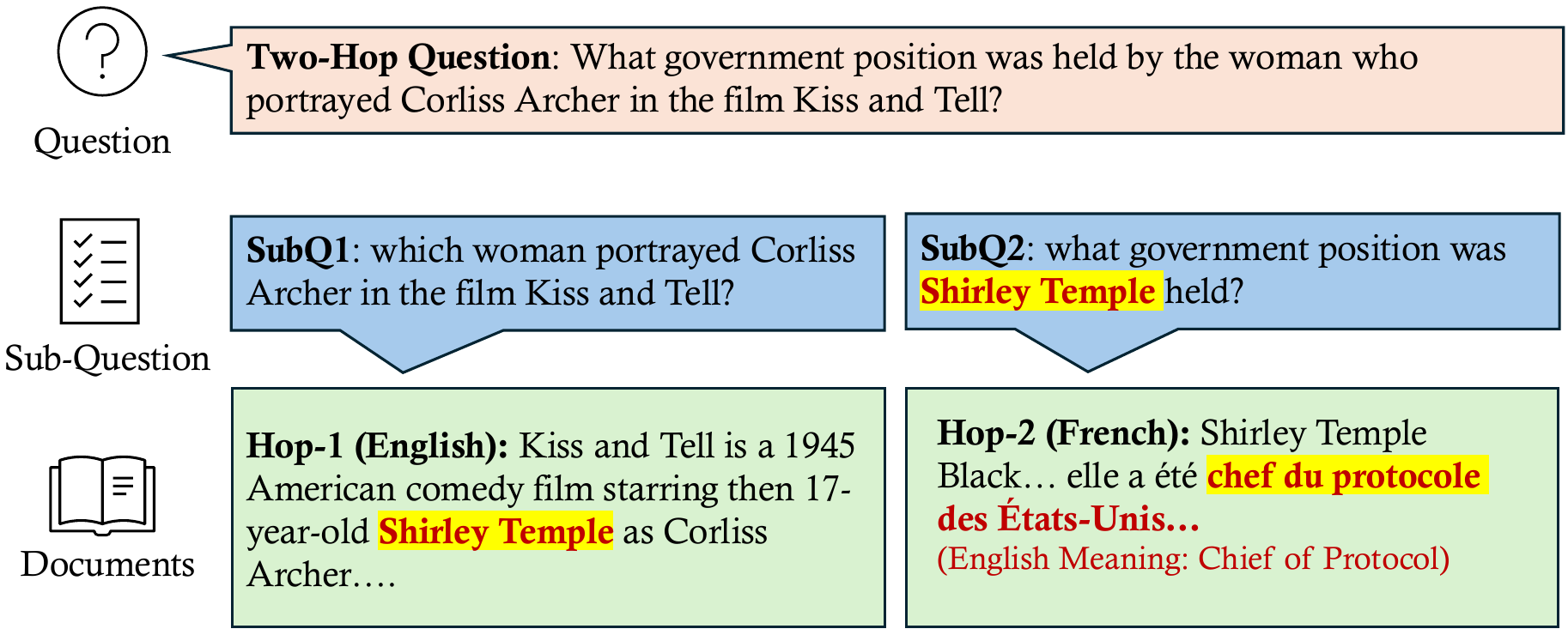}
        \caption{ Example of \textit{multilingual two-hop QA}. In Section \ref{sec:2}, we evaluate 
        multilingual two-step reasoning performance with the two-hop question and the corresponding \text{Hop-1} and \text{Hop-2} document. In Section \ref{sec:3}, we conduct a sub-question evaluation to disentangle the two-step reasoning mechanism: \text{SubQ1} infers the bridge entity and \text{SubQ2} links the bridge entity to the final answer. } 
        \label{fig:demo}
    \label{fig:fail-unfaith}
\end{figure*}

%To this end, we introduce a controlled multilingual two-hop question answering setting\footnote{We will release the dataset publicly to support reproducibility and further research.} 
%To this end, we introduce a controlled multilingual two-hop question answering setting\footnote{We will release the dataset publicly to support reproducibility and further research.} which bridges the gap between the progress on English-based two-hop reasoning and multilingual NLP. 
%To this end, we introduce a controlled multilingual two-hop question answering dataset\footnote{We will release the dataset publicly to support reproducibility and further research.} 

To this end, we extend two English-language multi-hop datasets, the decomposed version of \textsc{HotpotQA} \citep{tang-etal-2021-multi} and \textsc{MuSiQue} \citep{trivedi-etal-2022-musique}, to four diverse and high-resource languages: French, Russian, Arabic, and Chinese.\footnote{Dataset available at \url{https://github.com/vivian-my/X-Hop}.}
This extension broadens the reasoning evaluation across languages but also bridges the gap between English-centric two-hop reasoning and multilingual NLP.
Take Figure \ref{fig:demo} as an example.  
To answer a two-hop question: \emph{``What government position was held by the woman who portrayed Corliss Archer in the film Kiss and Tell?''}, it requires two reasoning steps to infer the final answer: 
First, identifying the bridge entity {Shirley Temple} as the woman who portrayed the film from an English first-hop document. 
Second, connecting {Shirley Temple} with the government position {chief of protocol} in a French second-hop document. 
This setup provides a clear testbed to examine whether models perform reasoning across contexts in varying languages and whether they reason in a step-by-step manner, commonly analogous to how humans solve problems via sub-questions.

From a controlled evaluation of multilingual large language models on this benchmark, we observe a degradation in two-hop QA performance when varying languages for each document (see Section~\ref{sec:2}).
In particular, changing the language of the answer-span (second-hop) documents leads to larger performance drops than changing the language of the bridging (first-hop) documents.
This discrepancy raises the question of whether the first step of multilingual reasoning is intrinsically easier than the second, or whether models rely on other mechanisms that bypass faithful reasoning.

Subsequently, we conduct a step-level evaluation on decomposed sub-questions (see Section~\ref{sec:3}). 
%First, we show that correct multilingual two-hop reasoning does not necessarily imply faithful step-by-step sub-question answering. 
We observe a notable unfaithfulness ratio (up to $33\%$) in multilingual scenarios where language models correctly answer a two-hop question but fail to answer the first step's sub-question. 
This indicates that successfully performing two-hop QA does not strictly require models to explicitly solve intermediate reasoning steps. 

To understand how models reach the correct answer without identifying explicit bridging information, we conduct a context attribution analysis. 
The results reveal that the bridging documents still play a critical role in answer generation, with at least one-third of the attribution scores. 
Moreover, the performance of unfaithful two-hop cases drops sharply when distractors with topically related bridging information are introduced.
This shows that language models perform multilingual two-hop reasoning, but this reasoning does not indicate explicit step-by-step sub-question answering.

On the other hand, we show that faithful step-by-step sub-question answering does not necessarily ensure correct multilingual two-hop reasoning.
Following \citet{press-etal-2023-measuring}, we refer to this phenomenon as compositional failure, and observe up to $18\%$ of cases where the model correctly answers each sub-question but fails to integrate the intermediate information into the final two-hop answer.
To resolve this issue, we examine the impact of presenting explicit reasoning information. 
We first show that standard chain-of-thought prompting \citep{Wei2022ChainOT}, which requests models to ``think step-by-step'', only partially alleviates compositional failure.
Building on the decomposed sub-questions, we introduce a three-stage \textsc{SubQ} prompting method, where each sub-question is explicitly provided to guide the final answer generation.
This approach substantially reduces compositional failures, improving averaged token-level F1 from $7.16\%$ to $39.77\%$. 
These results highlight question decomposition as a promising direction for enhancing multilingual multi-step reasoning.

\begin{table*}[!t]
  \centering
    \resizebox{\linewidth}{!}{%
      \begin{tabular}{lllcccccc}
        \toprule
    \multirow{2}{*}{\textbf{Settings}}   & \multicolumn{2}{c}{\textbf{Document Languages}} & \multicolumn{5}{c}{\textbf{Query Languages ($L_q$)}}  &  \multirow{2}{*}{\textbf{Avg.}}\\
        \cmidrule(lr){2-3} \cmidrule(lr){4-8}
       & \textbf{\text{Hop-1}} & \textbf{\text{Hop-2}} & \textbf{En} & \textbf{Fr} & \textbf{Ru} & \textbf{Ar} & \textbf{Zh}  & \\
        \midrule
%      \textbf{Monolingual} & $Q$ & $Q$ & 
      \textbf{Monolingual} & $=L_q$ & $=L_q$ & 
      \colorbox{e7}{56.08} & \colorbox{e10}{49.04} & \colorbox{e13}{43.48} & \colorbox{e13}{43.70} & \colorbox{e18}{32.25} & 44.91 \\
        \hdashline 
      \multirow{3}{*}{\textbf{Multilingual}} 
%      & $\overline{Q}$ & $Q$ 
      & $\ne L_q$ & $=L_q$ 

      & \colorbox{e7}{55.49} & \colorbox{e10}{49.19} & \colorbox{e13}{43.12} & \colorbox{e13}{42.92} & \colorbox{e17}{35.41} & 45.22 \\
%              & $Q$ & $\overline{Q}$ 
              & $=Lq$ & $\ne L_q$ 
              & \colorbox{e10}{48.58} & \colorbox{e12}{45.67} & \colorbox{e14}{40.01} & \colorbox{e14}{40.19} & \colorbox{e16}{36.13} & 42.12 \\
%              & $\overline{Q}$ & $\overline{Q}$ 
              & $\ne L_q$ & $\ne L_q$ 
              & \colorbox{e10}{48.16} & \colorbox{e12}{44.64} & \colorbox{e14}{41.37} & \colorbox{e14}{40.50} & \colorbox{e14}{40.33} & 43.00 \\
        \bottomrule
      \end{tabular}}
    \caption{Multilingual two-hop QA performance for Gemma-3-27B-Instruct. We report token-level F1 scores based on exact token matching. \mbox{$=L_q$} or \mbox{$\ne L_q$} denote whether the languages of documents are the same or different from the query language. For non-matching documents, we report the average across the four non-query languages. Full results are in Appendix \ref{app:full-multilingual-results}. Results for additional models, including Llama-3.1, Qwen-3, and Aya-Expanse, are reported in Appendix \ref{app:additional-models}.}
  \label{tab:combined-2hop-gemma}
\end{table*}

\section{Multilingual Two-Hop QA}  \label{sec:2}

The multilingual two-hop question answering (QA) task requires language models to reason with information from two gold documents in different languages to generate the answer. 
In this task, the models' inputs are: (i) a two-hop question to answer and (ii) two gold documents.
We denote \text{{Hop-1}} as first-hop documents containing bridge entities and \text{Hop-2} as second-hop documents with target answers. 
%that contain bridging (first-hop) and target answer (second-hop) information. 
In particular, a bridge entity must first be identified to infer the final answer in the second-hop documents.
This task evaluates models' multilingual understanding and reasoning ability to integrate information across different languages.

\subsection{Experimental Setup}
\paragraph{Datasets.}

We evaluate on the extended decomposed \textsc{HotpotQA} dataset~\citep{tang-etal-2021-multi}, derived from \citet{yang-etal-2018-hotpotqa}, which comprises $1{,}000$ English two-hop questions with corresponding decomposed sub-questions; 
we use \text{SubQ1} and \text{SubQ2} to denote the first and second single-hop sub-questions.
To probe multi-hop reasoning at varying depths, we additionally use \textsc{MuSiQue}~\citep{trivedi-etal-2022-musique}, whose dev set contains $2{,}417$ compositional questions spanning $2$-, $3$-, and $4$-hop reasoning chains.
  
\paragraph{Dataset Filter.}
To ensure models rely on true compositional reasoning rather than memorized pre-training knowledge, we filter both datasets as follows. 
For a $k$-hop question with $k$ gold documents $\{d_{\text{hop}_1}, \ldots, d_{\text{hop}_k}\}$, we discard any data instance that the experimental
models can answer correctly from any subset of these documents, i.e., any combination excluding at least one required document. 
This ensures that deriving the correct answer strictly requires the complete $k$-document chain.
After filtering, the evaluation corpus contains $1{,}312$ examples: $803$ two-hop, $327$ three-hop, and $182$ four-hop.

\paragraph{Dataset Translation.}
We automatically translate the filtered English datasets into four high-resource languages with varying language families and written scripts: French (Fr), Russian (Ru), Arabic (Ar), and Chinese (Zh).
%The translation prompts and quality evaluation are shown in Appendix x. 
We use \texttt{GPT-4o-mini} to translate the filtered English multi-hop datasets into four selected target languages.\footnote{Sampling parameters for translation: temperature = $1.0$ and Top-$p$ = $1.0$.} 
To ensure translation quality, we conducted human evaluations on a subset of translation examples (see Appendix \ref{app:translation-quality}).

\begin{table}[!t]
    \centering
    \small
    \renewcommand{\arraystretch}{1.1}
    \begin{tabular*}{\columnwidth}{@{\extracolsep{\fill}}lccc@{}}
        \toprule
        \textbf{Setting} & \multicolumn{3}{c}{\textbf{Number of Hops}} \\
        \cmidrule(lr){2-4}
                         & \textbf{2} & \textbf{3} & \textbf{4} \\
        \midrule
        Monolingual F1  & 56.1 & 48.8 & 40.5 \\
        Cross-lingual F1 & 50.8 & 43.2 & 30.9 \\
        \midrule
        $\Delta$ F1     & -5.3 & -5.6 & -9.6 \\
        \bottomrule
    \end{tabular*}
    \caption{Token-level F1 scores of {Gemma-3-27B-Instruct} as the number of
    reasoning hops increases. The query here is English. In the monolingual
    setting, all supporting documents are English. In the cross-lingual
    setting, at least one supporting document is not English (averaged over Chinese,
    French, Arabic, and Russian).}
    \label{tab:reasoning-hops}
\end{table}

\paragraph{Models.} \label{exp:models}
We experiment with the Gemma-3-Instruct \citep{Kamath2025Gemma3T} model of size $27$B due to its strong multilingual ability, which supports over $140$ languages. 
To show the generalisability of our findings, we report results from additional multilingual models, including Qwen-3-8B, Llama-3.1-8B-Instruct, and Aya-Expanse-8B, in Appendix \ref{app:additional-models}.
Following \citet{modarressi2025nolima}, we adopt greedy decoding for question answering tasks. 
For the prompts, we put the two-hop question \textit{before} and \textit{after} the provided documents to reduce the effect of query-aware contextualization \citep{Liu2023LostIT}.  
The standard prompt templates are in Appendix \ref{app:prompts}.

\begin{figure*}[t]
      \centering
      \includegraphics[width=\textwidth]{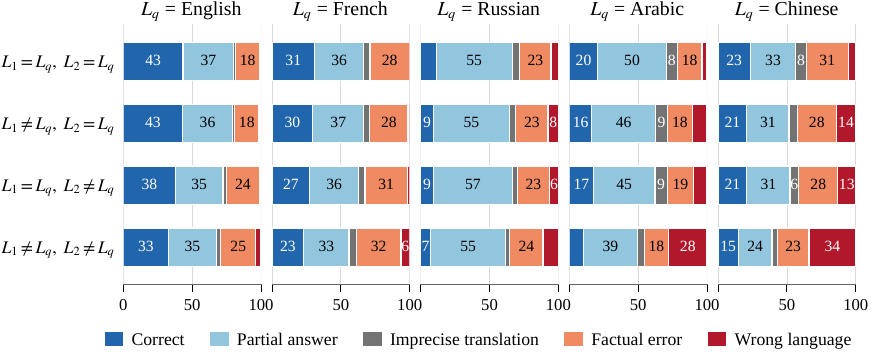}
      \caption{ Error analysis of two-hop QA predictions from Gemma-3-27B-Instruct.
$L_q$, $L_1$, and $L_2$ denote the languages of the query, Hop-1 document, and Hop-2 document, respectively.
}
     \label{fig:gemma-hop-language-alignment}
\end{figure*}

\subsection{Results}

\paragraph{Cross-lingual degradation persists across reasoning depths.}
Table~\ref{tab:combined-2hop-gemma} reports multilingual two-hop QA performance under different query languages. 
Across the evaluated multilingual language models, performance mostly decreases in cross-lingual settings where one or both document languages differ from the query language.
This degradation is not limited to two-hop reasoning: Table \ref{tab:reasoning-hops} shows that cross-lingual settings consistently underperform monolingual settings as the number of reasoning hops increases.

\paragraph{Reasoning-related errors dominate cross-lingual degradation.}
We define a fine-grained taxonomy of errors in two-hop QA predictions
(Appendix~\ref{sec:error-analysis}), including partial answers,
imprecise translations, factual errors, and wrong-language responses.
We annotate $500$ predictions from Gemma-3-27B for each query language ($2,500$ predictions in total) using \texttt{Claude-OPUS-4.6}.
Figure~\ref{fig:gemma-hop-language-alignment} shows that factual failures and imprecise answers are the major errors in both monolingual and cross-lingual settings. 
This suggests that the cross-lingual degradation primarily reflects failures in reasoning over the multilingual information, rather than errors in translating final answers. 
At the same time, the failure modes vary across query languages.
For English and French queries, changing to cross-lingual contexts mainly increases factual failures.
For Arabic, Chinese, and Russian queries, wrong language predictions become more frequent in cross-lingual settings, although reasoning-related errors remain the primary source.

\paragraph{Hop-2 language matters more than Hop-1.}
Table~\ref{tab:combined-2hop-gemma} shows that performance is substantially more sensitive to the language of the answer-span document (\text{Hop-2}) than to that of the bridging document (\text{Hop-1}). 
Across all models, changing the \text{Hop-1} language yields only a marginal average change relative to the monolingual setting ($-0.83$ F1), while changing the \text{Hop-2} language leads to a much larger average drop of $-8.49$ F1. This pattern holds across all models, although the magnitude varies considerably, from $-2.79$ F1 for Gemma-3-27B to $-13.43$ F1 for Aya-Expanse-8B.
To examine whether this degradation relates to linguistic similarity, we compute Pearson and Spearman correlations between QA performance and the linguistic similarity between the question and each document.\footnote{Details of the linguistic similarity calculation are provided
in Appendix \ref{app:linguistic-similarity}} 
Table \ref{tab:multi-hop-corr} shows that final performance has a strong correlation with the linguistic similarity between the question and the Hop-2 document.

\begin{table*}[!t]
\centering
\resizebox{0.95\linewidth}{!}{%
\begin{tabular}{lcccccccccc}
\toprule
\multirow{2}{*}{\textbf{Correlation}} 
& \multicolumn{2}{c}{\textbf{En}} 
& \multicolumn{2}{c}{\textbf{Fr}} 
& \multicolumn{2}{c}{\textbf{Ru}} 
& \multicolumn{2}{c}{\textbf{Ar}} 
& \multicolumn{2}{c}{\textbf{Zh}} \\
\cmidrule(lr){2-3} \cmidrule(lr){4-5} \cmidrule(lr){6-7} \cmidrule(lr){8-9} \cmidrule(lr){10-11}
& \textbf{Hop-1} & \textbf{Hop-2} 
& \textbf{Hop-1} & \textbf{Hop-2} 
& \textbf{Hop-1} & \textbf{Hop-2} 
& \textbf{Hop-1} & \textbf{Hop-2} 
& \textbf{Hop-1} & \textbf{Hop-2} \\
\midrule
\textbf{Pearson}   & \colorbox{r18}{0.01} & \colorbox{r10}{0.89} & \colorbox{r17}{0.06} & \colorbox{r12}{0.66} & \colorbox{r15}{0.20} & \colorbox{r12}{0.78} & \colorbox{r15}{0.20} & \colorbox{r11}{0.70} & \colorbox{r16}{0.12} & \colorbox{r12}{0.69} \\
\textbf{Spearman}  & \colorbox{r17}{0.09} & \colorbox{r14}{0.70} & \colorbox{r18}{0.01} & \colorbox{r12}{0.60} & \colorbox{r17}{0.09} & \colorbox{r12}{0.60} & \colorbox{r16}{0.12} & \colorbox{r13}{0.50} & \colorbox{r18}{0.01} & \colorbox{r13}{0.50} \\
\bottomrule
\end{tabular}
}
\caption{{Pearson/Spearman correlations between two-hop QA performance and the linguistic similarity for Gemma-3-27B-Instruct. Overall, linguistic similarity between two-hop questions and \text{Hop-2} documents has a strong correlation with the performance.}}
\label{tab:multi-hop-corr}
\end{table*}

\section{Multilingual Reasoning Decomposition} \label{sec:3}

In this section, we disentangle the two-hop reasoning process by two single steps through our translated subset of decomposed sub-questions from \textsc{HotpotQA}. 
This explicit step-wise evaluation facilitates a more fine-grained understanding of multilingual two-hop reasoning behavior.

\subsection{Setup}

We denote \text{SubQ1} as the first-step sub-questions that extract bridge entities from \text{Hop-1} documents, and \text{SubQ2} as the second-step sub-questions that retrieve final answers from \text{Hop-2} documents; see also Figure~\ref{fig:demo}. 
Based on the decomposed evaluation, we examine whether multilingual multi-hop reasoning follows a faithful step-by-step decomposition. 
We further identify two distinct failure modes following the definition in previous works, i.e., \emph{unfaithfulness} \citep{lyu-etal-2023-faithful} and \emph{compositional failure} \citep{press-etal-2023-measuring}.

%we identify two distinct modes of step-by-step sub-question answering failures, illustrated in Appendix Figure \ref{fig:demo_failuer}.
%From the evaluation of two-hop questions and their corresponding sub-questions, we identify two distinct modes of reasoning failures, illustrated in Appendix Figure \ref{fig:demo_failuer}.
%In this section, we analyze these two failure modes and discuss their potential causes.

\paragraph{Unfaithfulness.} Models correctly answer a two-hop question while failing to answer its sub-questions, i.e., \text{SubQ1} or \text{SubQ2}. The unfaithfulness ratio is calculated as the percentage of total unfaithful cases over all correctly answered two-hop questions. This ratio reflects whether models are faithful to {step-by-step} sub-question reasoning. 
   
\paragraph{Compositional Failure.} Models incorrectly answer two-hop questions while succeeding in both \text{SubQ1} and \text{SubQ2}. The composition failure ratio is calculated as the percentage of total composition failure cases over all incorrectly answered two-hop questions. This ratio reflects the limitation of language models' compositional reasoning.

\begin{table*}[!t]
    \centering
    \begin{subtable}[t]{\linewidth}
        \centering
    \resizebox{\linewidth}{!}{%
    \begin{tabular}{  p{1.8cm} p{1.7cm}  % col 2: Multi-hopQ
    p{1.5cm}  % col 3: Sub1Q
    p{1.5cm}  % col 4: Sub2Q
    p{1.0cm}  % col 5: En
    p{1.0cm}  % col 6: Fr
    p{1.0cm}  % col 7: Ru
    p{1.0cm}  % col 8: Ar
    p{1.0cm} }
        \toprule
      \multirow{2}{*}{\textbf{Setting}} & \multicolumn{3}{c}{\textbf{Answer Correctness}}   & \multicolumn{5}{c}{\textbf{Query Languages}}  \\
       \cmidrule(lr){2-4} \cmidrule(lr){5-9}
    &  {\textbf{\text{Two-HopQ}}} &  {\textbf{\text{SubQ1}}} & {\textbf{\text{SubQ2}}} &   \textbf{En} & \textbf{Fr} & \textbf{Ru} & \textbf{Ar} & \textbf{Zh} \\
    \midrule
   \multirow{3}{*}{{\textbf{Monolingual}}} &  \cmark & \xmark  & \cmark  & \colorbox{e16}{0.07} & \colorbox{e12}{0.19} & \colorbox{e16}{0.12} & \colorbox{e16}{0.10} & \colorbox{e15}{0.16}  \\
    & \cmark & \cmark  & \xmark  & \colorbox{e19}{0.02} & \colorbox{e20}{0.00} & \colorbox{e20}{0.00} & \colorbox{e19}{0.05} & \colorbox{e19}{0.04} \\ 
     & \cmark & \xmark  & \xmark  & \colorbox{e19}{0.03} & \colorbox{e20}{0.00} & \colorbox{e20}{0.02} & \colorbox{e20}{0.00} & \colorbox{e20}{0.00} \\ 
    \hdashline
    \multirow{3}{*}{{\textbf{Multilingual}}} &  \cmark & \xmark  & \cmark &  \colorbox{e16}{0.12} & \colorbox{e11}{0.23} & \colorbox{e15}{0.15} & \colorbox{e11}{0.25} & \colorbox{e10}{0.33} \\ 
    & \cmark & \cmark  & \xmark & \colorbox{e19}{0.03} & \colorbox{e18}{0.03} & \colorbox{e19}{0.03} & \colorbox{e18}{0.05} & \colorbox{e18}{0.04} \\
    & \cmark & \xmark  & \xmark & \colorbox{e19}{0.01} & \colorbox{e20}{0.00} & \colorbox{e19}{0.03} & \colorbox{e19}{0.03} & \colorbox{e19}{0.02} \\

    \bottomrule
    \end{tabular}}
    \caption{Unfaithfulness Ratio}
     \label{tab:unfaith}
    \end{subtable}
    \vspace{0.2cm}

\begin{subtable}[t]{\linewidth}
        \centering
 \resizebox{\linewidth}{!}{%
    \begin{tabular}{  p{1.8cm}  % col 1: Setting
    p{1.7cm}  % col 2: Multi-hopQ
    p{1.5cm}  % col 3: Sub1Q
    p{1.5cm}  % col 4: Sub2Q
    p{1.0cm}  % col 5: En
    p{1.0cm}  % col 6: Fr
    p{1.0cm}  % col 7: Ru
    p{1.0cm}  % col 8: Ar
    p{1.0cm} }
        \toprule
     {\textbf{Monolingual}} & \xmark & \cmark  & \cmark  & \colorbox{e14}{0.14} & \colorbox{e13}{0.16} & \colorbox{e19}{0.03} & \colorbox{e13}{0.17} & \colorbox{e13}{0.18} \\ 
     \hdashline
    {\textbf{Multilingual}}  & \xmark & \cmark & \cmark & \colorbox{e16}{0.11} & \colorbox{e16}{0.10} & \colorbox{e18}{0.05} & \colorbox{e16}{0.10} & \colorbox{e17}{0.08} \\
    
    \bottomrule
    \end{tabular}}
    
    \caption{Compositional Failure Ratio}
    \label{tab:failure}
\end{subtable}
    \caption{ The decomposed sub-questions evaluation for Gemma-3-27B-Instruct. We report the average ratio for multilingual settings.  }
     \label{tab:decompose}
\end{table*}

%\begin{figure*}[!t]
%    \centering
%    \includegraphics[width=\linewidth]{iclr2026/images/intro/failure.pdf}
% \caption{ Two distinct reasoning failure modes from Gemma-3-27B-Instruct. \textbf{Left}: Unfaithfulness, \textbf{Right}: Composition Failure. The full documents for \text{Hop-1} and \text{Hop-2} are shown in Appendix x. }
%    \label{fig:demo_failuer}
%\end{figure*}

\subsection{Unfaithfulness} \label{sec:3.1}

\label{overall-results}

%Table \ref{tab:unfaith} reports the unfaithfulness ratio for both monolin
%, defined as the proportion of failed sub-questions among correctly answered two-hop questions. 
Table \ref{tab:unfaith} shows that the model is more likely to be unfaithful to the first-step sub-question for both monolingual and multilingual settings. 
Consistent with \citet{tang-etal-2021-multi}, we show that explicitly identifying the bridge entity is not required for the model to answer two-hop questions correctly in monolingual English settings. 
Furthermore, we find that multilingual settings yield higher unfaithfulness ratios than monolingual ones.
%This shows that the model is more likely to be unfaithful to the first sub-question when the documents are multilingual. 
In particular, Arabic and Chinese two-hop questions yield notably high first-step unfaithfulness rates of $25\%$ and $33\%$, respectively. 
%These results show that models do not perform multi-hop reasoning by answering an intermediate sub-question to infer the final answer. 
To further probe how language models are still able to generate correct answers in unfaithful cases, we conduct a context utilization analysis to examine the role of both hop documents, especially for bridging documents.

\subsubsection{Context Attribution Analysis}
Here, we aim to implicitly analyze how language models use different hop documents to generate the answers. 
We follow input attribution methods to measure how much each document contributes to generating the answers \citep{DBLP:conf/nips/LundbergL17,DBLP:journals/jmlr/CovertLL21}. 

\paragraph{Setup.} We use the input-erasure attribution method \citep{DBLP:journals/corr/LiMJ16a} to measure the contribution of each hop's document on the correct answer generation. 
Input erasure quantifies the contribution of an input component (typically a token) by measuring the change in the model’s probability on the ground truth when that component is removed.
In our experiments, we compute token-level attributions with respect to producing the correct answer. 
A document’s attribution is obtained by summing the contributions of its tokens, after which we calculate its percentage attribution relative to the combined attribution of the two documents, i.e., \text{Hop-1} and \text{Hop-2}. 
Finally, we report the average percentage attribution for each document across the entire data. 
To ensure a fair comparison, we examine the context attribution scores in unfaithful cases on the first sub-question against fully faithful ones\footnote{Two-hop, SubQ1, and SubQ2 all correct} to show shifts in context attribution.

\begin{figure}[!t]
    \centering
   
    \includegraphics[width=\linewidth]{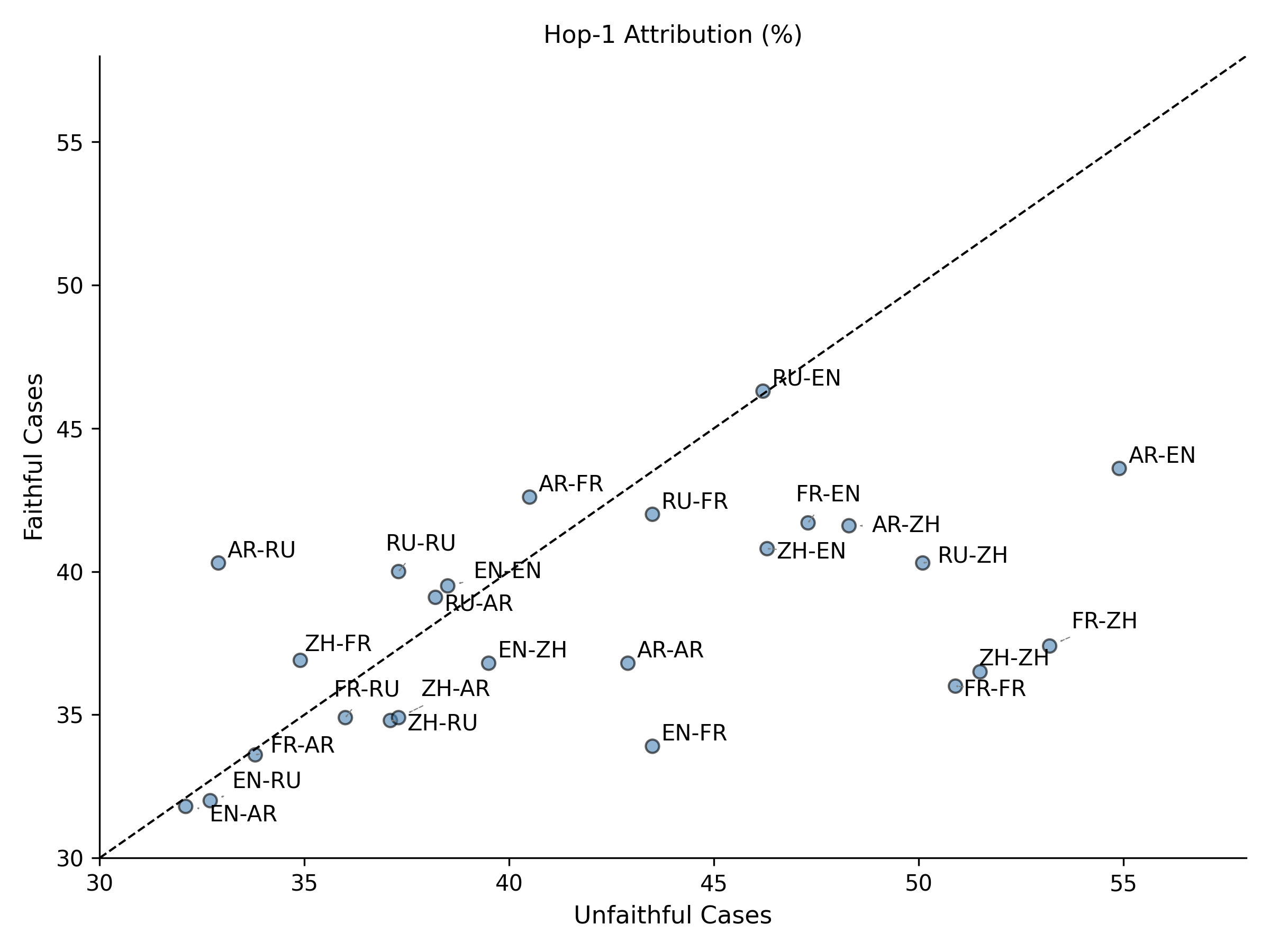}
    \caption{Context attribution scores for faithful and unfaithful cases. %Hop-1 attribution percentages are computed relative to the combined contributions of Hop-1 and Hop-2 documents. 
    The two-hop query is in English, and Lang1-Lang2 (e.g., EN-ZH) indicates the languages of Hop-1 and Hop-2 documents.}
    \label{fig:context_attribution}
\end{figure}

\paragraph{Results.}
Figure \ref{fig:context_attribution} presents \text{Hop-1} attribution scores for faithful and unfaithful cases in English two-hop questions. 
In general, we surprisingly notice that most unfaithful cases exhibit relatively higher \text{Hop-1} attribution scores compared to faithful cases.  
Although the model fails to explicitly infer bridge entities from \text{Hop-1} documents for unfaithful cases, \text{Hop-1} documents still implicitly contribute to the correct two-hop target answer generation. 
The relatively high \text{Hop-1} attribution scores for unfaithful cases indicate the model relies on a more nuanced intrinsic reasoning mechanism that extends beyond what can be revealed through explicit step-level evaluation.
Moreover, we notice that multilingual \text{Hop-1} documents have higher attributions than English ones for both cases.
Taken together with the observation in \Cref{overall-results} that multilingual settings exhibit higher unfaithfulness, this suggests that while the models attend to the multilingual \text{Hop-1} documents, they struggle to answer the sub-question on them correctly.

%In general, we notice that multilingual \text{Hop-1} documents have higher attributions than English ones for both cases. 
%Taken together with the observation in \Cref{overall-results} that multilingual settings exhibit higher unfaithfulness, this suggests that while the models attend to the multilingual \text{Hop-1} documents, they struggle to interpret them correctly.

%Furthermore, we observe that most unfaithful cases exhibit relatively higher \text{Hop-1} attribution scores compared to faithful cases. 
%Although these correctly answered unfaithful cases fail on the first sub-questions, \text{Hop-1} documents still implicitly contribute to answer generation. 
%This highlights the complexity of LLMs' implicit reasoning, which cannot be fully captured using a disentangled step-by-step sub-question reasoning. 

%We aim to examine how the model generates correct answers in using \text{Hop-1} or \text{Hop-2} context. 
%First, we show that the model still requires first-hop documents to generate correct answers, where the scores are xx. 
%Moreover, the multilingual documents have higher attribution scores compared to monolingual ones. 

\begin{figure*}[!t]
    \centering
    \begin{subfigure}[b]{0.49\textwidth}
        \centering
        \includegraphics[width=\linewidth]{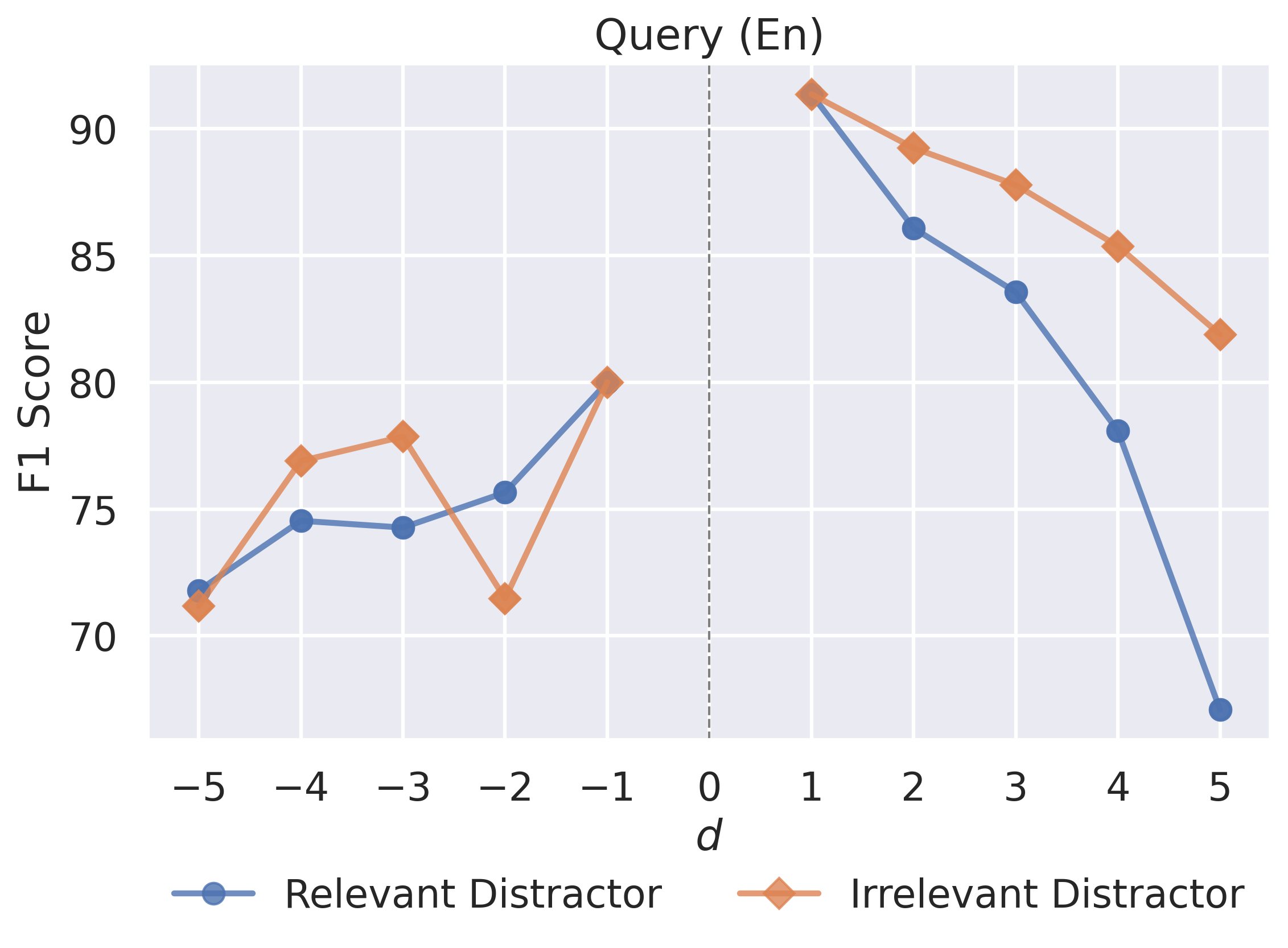}
     
    \end{subfigure}
    \hfill 
    \begin{subfigure}[b]{0.49\textwidth}
        \centering
        \includegraphics[width=\linewidth]{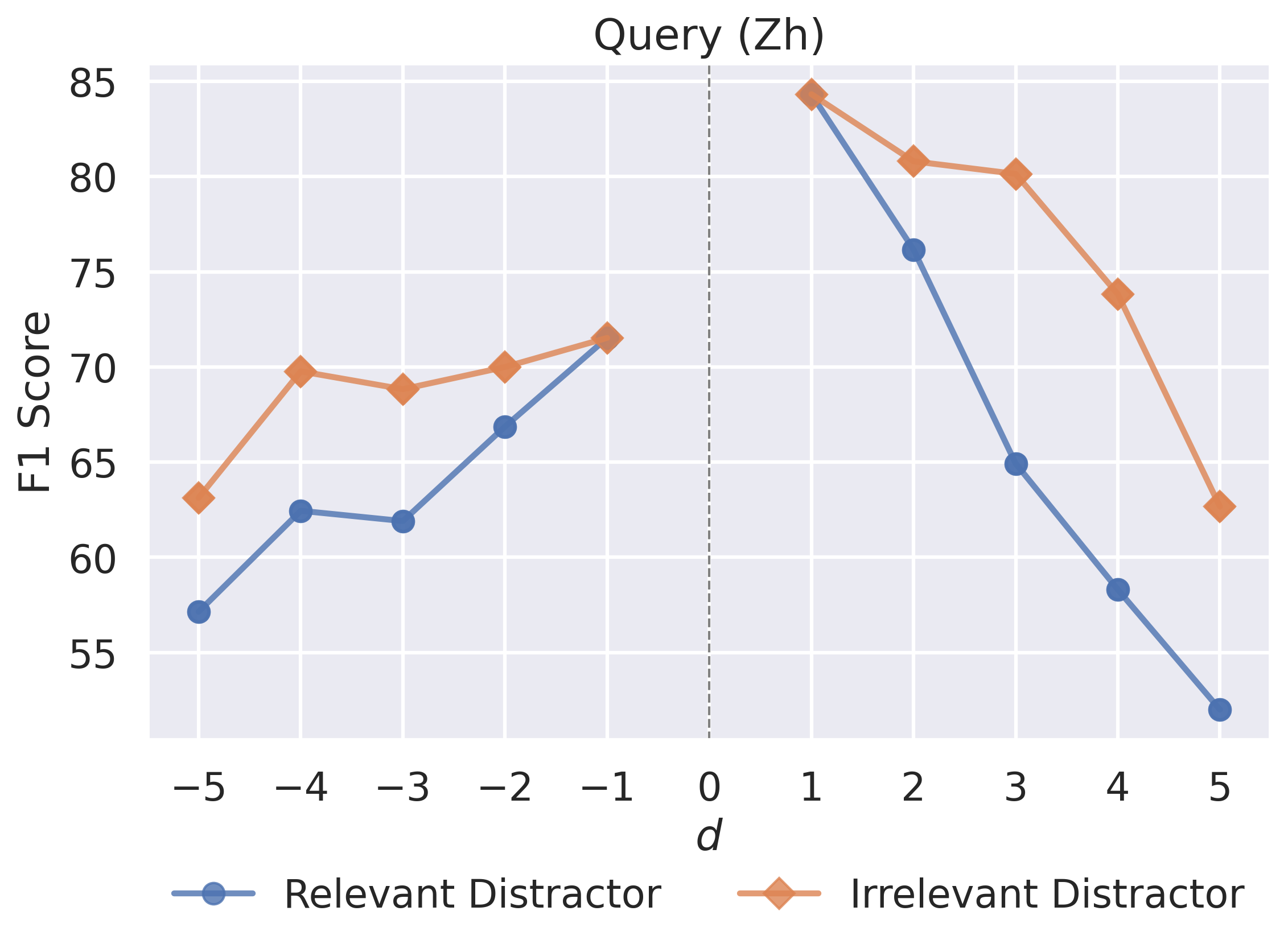}
     
    \end{subfigure}
    \caption{ The impact of inserting relevant and irrelevant distractors between \text{Hop-1} and \text{Hop-2} documents. A distance of $d$ corresponds to $(|d|-1)$ distractors between the two hops. Positive $d$ means \text{Hop-1} precedes \text{Hop-2}, while a negative sign means the reverse. We report the average token-level F1 scores for every unfaithful multilingual case for each query language.}
    \label{fig:distractor}
\end{figure*}

\subsubsection{Context Distractor Analysis}

The context attribution analysis shows the importance of \text{Hop-1} documents on intrinsic multilingual reasoning for unfaithful cases.
To further investigate whether the reliance on \text{Hop-1} documents reflects robust reasoning rather than shortcutting spurious cues, we perform a controlled context perturbation analysis on unfaithful cases inspired by prior works by distractors \citep{hengle-etal-2025-multilingual} and context orders \citep{yu-etal-2025-unleashing}. 

%We consider two factors: 
%(1) Topic-controlled distractors: we introduce distractor documents with controlled topic-wise relevance to the \text{Hop-1} documents. 
%The sensitivity of the LLM's performance on varying topic-relevant distractors reflects the robust reasoning. 
%(2) Context order: 
%In specific, we introduce distractor documents with controlled topic-wise relevance to the \text{Hop-1} documents to examine the LLM's ability to distinguish true bridging information. 
%Furthermore, we reverse the order of with \text{Hop-1} and \text{Hop-2} documents to 
%To further examine the role of \text{Hop-1} documents on multilingual two-hop reasoning, we introduce distractor documents with controlled topic-wise relevance to the \text{Hop-1} documents. 

\paragraph{Setup.} 
We gradually insert an increasing number of distractor documents between \text{Hop-1} and \text{Hop-2} documents. 
The language of the distractors is the same as the two-hop question. 
We use distance $d$ to denote the position differences between \text{Hop-1} and \text{Hop-2}, and it corresponds to $(|d|-1)$ distractors between the two hops. 
The sign of $d$ specifies their order: a positive value means \text{Hop-1} precedes \text{Hop-2}, while a negative value means the reverse.
In particular, we control distractor relevance with the {bridging} \text{Hop-1} documents. 
Relevant distractors contain topics similar to the original bridging \text{Hop-1} documents from the dataset, whereas irrelevant distractors are randomly sampled from unrelated training examples.

\paragraph{Results.}

Figure~\ref{fig:distractor} shows that inserting distractors between \text{Hop-1} and \text{Hop-2} degrades multilingual two-hop reasoning, consistent with monolingual findings by \citet{modarressi2025nolima}. 
Topic-relevant distractors result in larger drops than irrelevant ones, underscoring the difficulty of models to discriminate true bridging evidence from semantically proximate noise and highlighting the central role of \text{Hop-1} in two-step reasoning.
Moreover, changing the document order further degrades performance: presenting the \text{Hop-1} and \text{Hop-2} documents in reverse order is consistently worse than the original sequence, echoing premise-order effects in LLMs \citep{Chen2024PremiseOM, yu-etal-2025-unleashing}. 
This indicates a positional sensitivity of intermediate bridge information, suggesting that the model relies on order-dependent reasoning rather than fully order-robust multi-hop composition.

\subsection{{Compositional Failure}}

%\subsubsection{Overall Results}

The previous section \ref{sec:3.1} analyzes the unfaithfulness cases and shows that correct intrinsic reasoning does not necessarily follow faithful step-by-step sub-question answering. 
Here, we examine the second failure mode, i.e., compositional failure, where the model answers sub-questions correctly but fails to answer the overall two-hop question. 
Table~\ref{tab:failure} shows compositional failure rates up to $18\%$, showing that solving both sub-questions does not guarantee correct two-hop composition. 
Motivated by evidence that prompting with explicit intermediate reasoning can improve performance \citep{Wei2022ChainOT}, we aim to examine whether guiding language models with (i) self-generated chain-of-thoughts (CoT) or (ii) decomposed step-by-step sub-questions, can reduce compositional failure.

\begin{figure}[!t]
    \centering
        \includegraphics[width=0.45\textwidth]{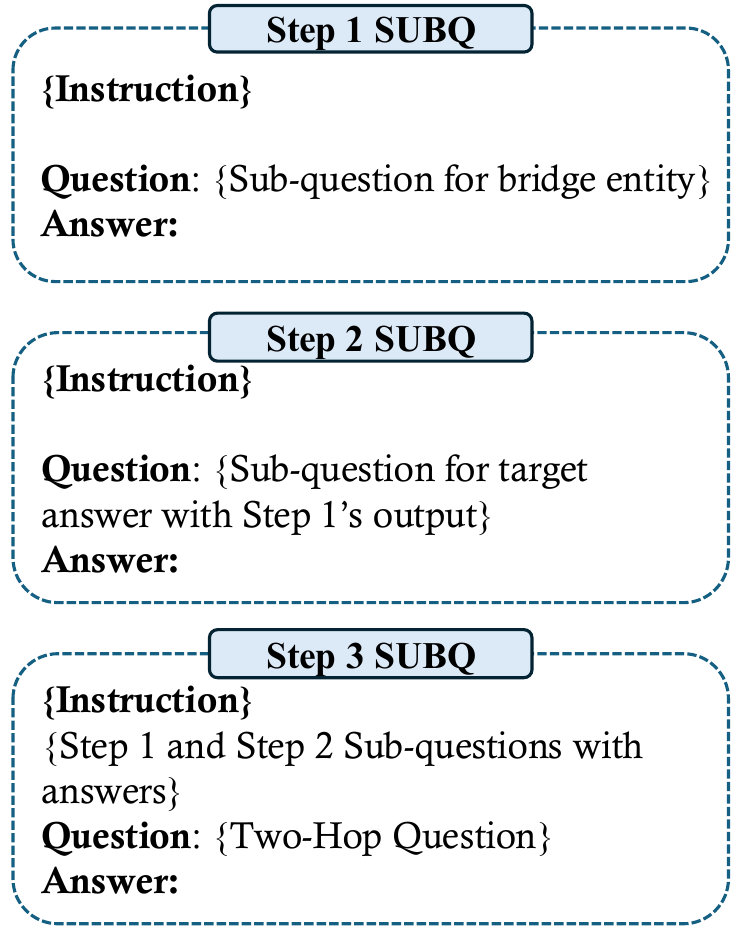}
        \caption{Three-step of \textsc{SubQ} Prompting. } %The first and second-step prompting templates are shown in x. }
    \label{fig:three_stage_prompt}
\end{figure}

%We consider two ways of incorporating intermediate reasoning information: (a) zero-shot CoT: model-generated reasoning chains and (b) SubQ Prompt: guided with decomposed sub-questions.
%This comparison allows us to assess the effectiveness of intermediate reasoning steps generated by the model itself versus those guided by structured decomposition. 
%The setup details and results are shown below. 

%training or prompting schemes that couple sub-tasks or sub-questions with generation. 
%self-generated chains may diverge from the task’s true decomposition, introducing errors in intermediate steps and increasing the likelihood of incorrect final outputs.

\begin{figure*}[!t]
    \centering
    \begin{subfigure}[b]{0.49\textwidth}
        \centering
        \includegraphics[width=\linewidth]{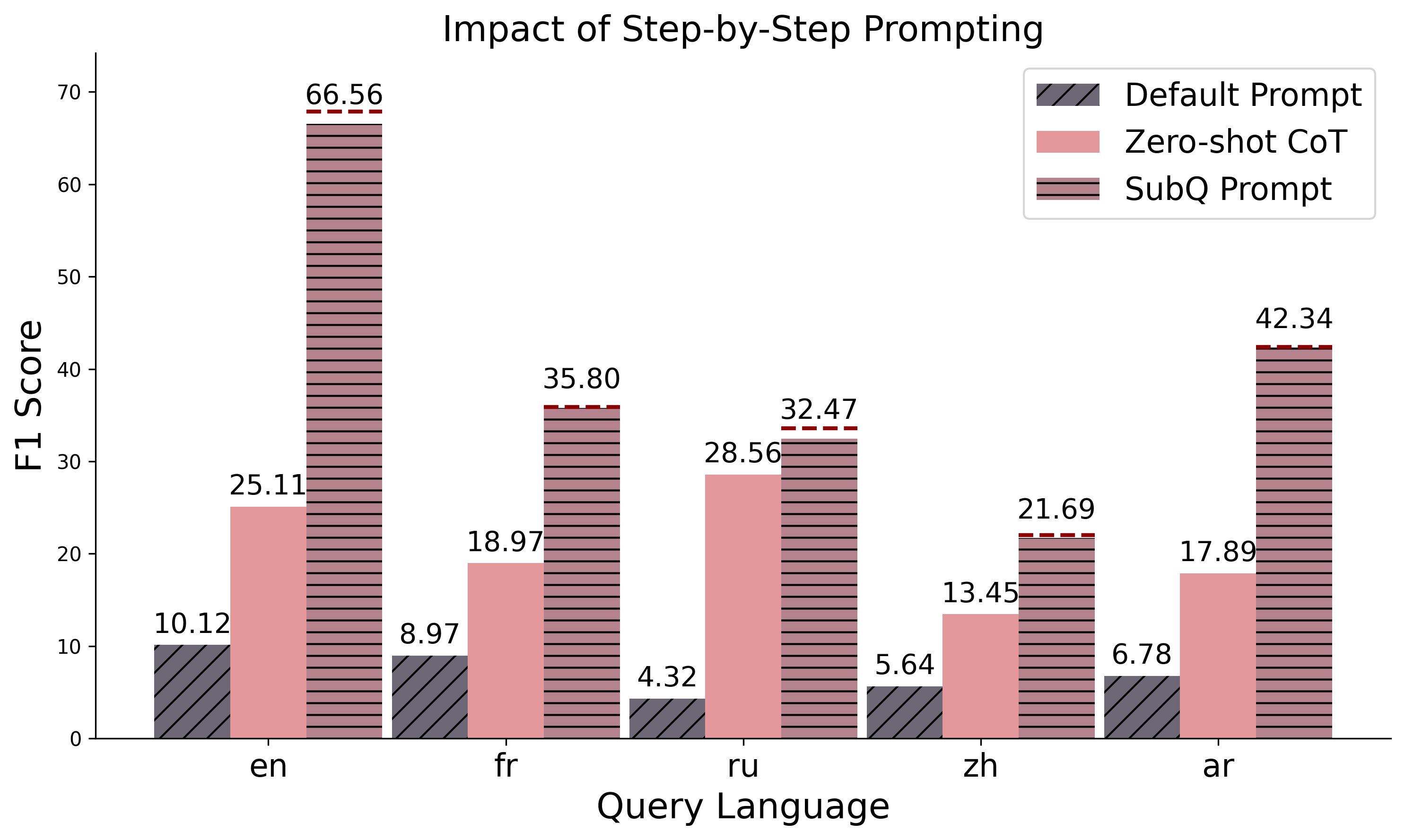}
        \caption{Composition Failure Cases}
        \label{fig:sub1}
    \end{subfigure}
    \hfill 
    \begin{subfigure}[b]{0.49\textwidth}
        \centering
        \includegraphics[width=\linewidth]{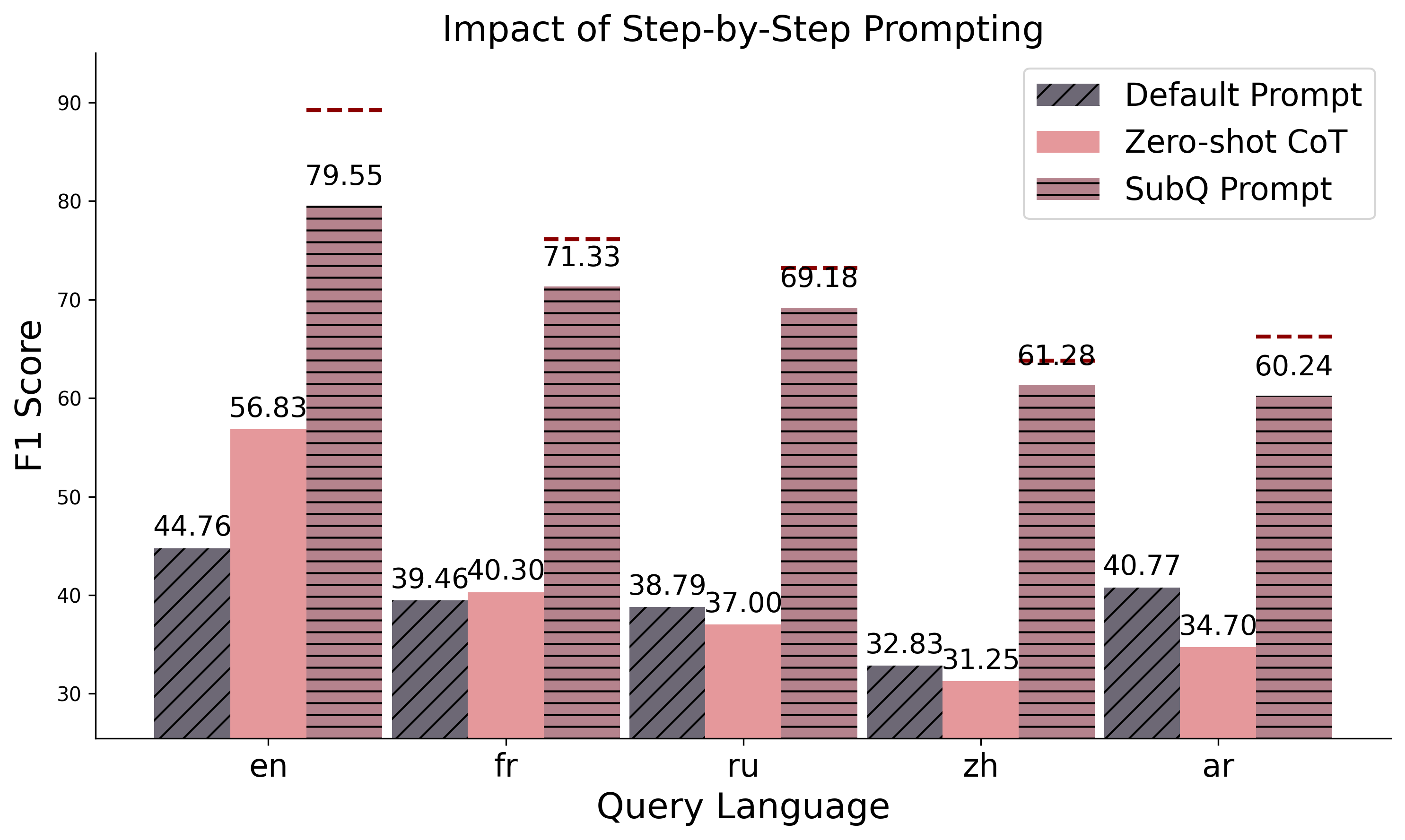}
        \caption{Full Set}
        \label{fig:sub2}
        
    \end{subfigure}
    
    \caption{ Token-level F1 scores for multilingual two-hop QA for different prompting strategies, evaluated on composition failure cases and the full dataset. \textcolor{red}{Red} dashed lines above the \text{SubQ} Prompt bars show the performance when using ground-truth answers for sub-questions. }
    \label{fig:prompt_diff}
\end{figure*}

\paragraph{Zero-shot CoT.} 
Following \citet{Kojima2022LargeLM}, we instruct the model to ``\textit{Think Step-by-Step}'' and generate its own reasoning chains for answering two-hop questions. 
Prompt templates are in Appendix Figure~\ref{fig:zero-shot-cot}.
The language of zero-shot instruction is aligned with the two-hop question, and we do not control the reasoning language here.

\paragraph{\textsc{SubQ} Prompt.}
We introduce a three-stage, step-by-step prompting technique with decomposed sub-questions. 
As illustrated in Figure~\ref{fig:three_stage_prompt}, prompting consists of three steps: First step, prompting the model to answer the \text{SubQ1} question about the bridge entity; Second step, using the first step output for the bridge entity inserted with \text{SubQ2} to ask for the target answer; 
Third step, presenting both sub-questions and the previous steps' outputs together to ask for the final two-hop answer.
For a fair comparison, we also use the ground-truth answers for each step's sub-question to guide reasoning.

Notably, \textsc{SubQ} is not intended as a deployable method, as it builds on gold decomposed sub-questions. 
Instead, it serves as a diagnostic upper bound for estimating how much cross-lingual degradation can be attributed to failures in question decomposition and intermediate reasoning.
Future work can investigate automatic decomposition methods to separate from the oracle supervision.

\subsubsection{Results.}

Figure \ref{fig:prompt_diff} shows the average multilingual two-hop QA performance for each query language with different prompting techniques. 
First, \textsc{SubQ} prompting yields substantial gains over zero-shot CoT on both the compositional failure cases and the full evaluation set.
These improvements suggest that explicitly decomposing the query into sub-questions helps the model integrate information step-wise.
One plausible explanation for the weaker zero-shot CoT results is that the model uses its own internal representations to generate explanations, which might diverge from the task's true decomposition. 
This, in turn, leads to errors in the reasoning.

Second, \textsc{SubQ} prompting shows a larger performance gap between model-generated predictions and ground-truth answers for sub-questions on the full evaluation set compared to the compositional failure cases.
This is expected, as the model predicts sub-questions correctly in compositional failure cases, resulting in performance close to that of the ground truth.
For the full set, despite potential errors in predicted answers of sub-questions, \textsc{SubQ} prompting still leads to substantial performance gains across all query languages.
%Overall, these results highlight the effectiveness of question decomposition in enhancing LLM performance on multilingual multi-hop reasoning tasks.

\section{Related Work}

In this section, we review prior work relevant to our study, focusing on multilingual capability and reasoning capability. 
\paragraph{Multilingual Capabilities.}
Multilingual capabilities have been extensively studied across downstream tasks, including machine translation \citep{stap-etal-2023-viewing, meng-monz-2024-disentangling}, question answering \citep{ponti2020xcopa, hendrycks2021measuring}, and mathematical reasoning \citep{shi2023language, qin-etal-2023-cross, chen2023breaking, Ghosh2025TheMM}.
Among these, multilingual mathematical reasoning is closely related to our study, but primarily examines reasoning over numerical operations rather than linguistic information.
Prior work has also used linguistic benchmarks, such as XCOPA \citep{ponti2020xcopa} and translated versions of MMLU \citep{hendrycks2021measuring}, to evaluate multilingual reasoning \citep{Chua2024CrosslingualCA}.
However, XCOPA typically requires only a single inference step, while MMLU does not provide explicit control over the intermediate steps involved in reaching an answer.
In contrast, we study multilingual multi-hop question answering in a controlled setting, disentangling the language used at each reasoning step to directly examine models' ability to integrate linguistic information across languages.

\paragraph{Intrinsic Reasoning.}
Intrinsic reasoning investigates whether LLMs are capable of reasoning without explicit prompting \citep{Wang2024ChainofThoughtRW}.
For instance, \citet{yang-etal-2024-large-language-models} reveals latent two-hop reasoning paths by constructing questions that require retrieving factual bridging information from pre-trained knowledge.
Similarly, \citet{Guo2025HowDL} trains a three-layer transformer model from scratch to examine how two-hop reasoning occurs by analyzing attention logits for bridge and target entities.
In contrast, our work evaluates reasoning explicitly through decomposed sub-questions, allowing us to assess whether multilingual reasoning aligns with a step-by-step sub-question answering process.

\paragraph{Failure Modes of Reasoning.}
This work is also related to several modes of reasoning failure in LLMs, including unfaithfulness \citep{lyu-etal-2023-faithful, Arcuschin2025ChainofThoughtRI}, premise order effects \citep{Chen2024PremiseOM, yu-etal-2025-unleashing}, distractibility \citep{Guo2025HowDL, Qi2025OnTC}, and the limited capacity for long-context reasoning \citep{Liu2023LostIT, modarressi2025nolima, hengle-etal-2025-multilingual}. 
Specifically, \citet{lyu-etal-2023-faithful} shows that chain-of-thought prompting on mathematical tasks often produces answers that do not faithfully follow the intermediate reasoning steps. 
Our work extends to multi-hop reasoning by showing the unfaithful outputs for decomposed sub-questions. 
\citet{modarressi2025nolima} reveals that increasing the context length of the two-hop associated reasoning QA task leads to performance degradation, while we show that this difficulty is further amplified when long-context reasoning is combined with multilinguality, exposing another limitation of current LLMs.

\paragraph{Effectiveness of Reasoning Decomposition.} 
Reasoning decomposition has been widely adopted in prompting techniques across various downstream tasks.
For example, chain-of-thought prompting decomposes mathematical problems into intermediate steps, encouraging models to follow a step-by-step reasoning strategy \citep{Wei2022ChainOT}.
In machine translation, \citet{briakou-etal-2024-translating, he-etal-2024-exploring} decompose the task into multiple stages, showing that multi-turn refinements can improve translation quality.
\citet{tang-etal-2021-multi} used decomposed sub-questions to evaluate multi-hop QA, and we extend this evaluation to cross-lingual settings. 
%While the decomposed sub-questions need to be additionally acquired, they provide a potential way of enhancing reasoning across languages.

\section{Conclusion}

In this paper, we introduce a controlled setting for multilingual two-hop reasoning to broaden multilingual evaluation. 
Building on this task, we present a comprehensive analysis that exposes the limitations of current strong multilingual language models in two-step reasoning when information is distributed across languages. 
Specifically, we find that language models are more sensitive to cross-lingual variation in answer-span documents than in bridging documents, despite both being equally essential for deriving the final answer.
Through both explicit and implicit analyses, we show that models rely on more nuanced intrinsic reasoning mechanisms that extend beyond what can be revealed by explicit step-by-step evaluation. 
On the other hand, this work demonstrates the benefits of explicit step-by-step prompting with structured sub-questions in multilingual multi-hop tasks, particularly for mitigating compositional failures. 
Although the model is capable of reasoning without predefined structures, our findings show that such guidance enhances its robustness.
These insights open promising directions for future work on incorporating reasoning decomposition into both training objectives and prompting strategies to improve multilingual multi-hop reasoning.

%By decomposing each query into sub-questions, we disentangle the two-step reasoning process and observe that models frequently fail the first sub-question (identifying the bridge entity) while still answering the overall two-hop question correctly.
%To probe unfaithfulness, we conduct context attribution analysis, which reveals that bridging information still plays a crucial role. 
%Our findings indicate that multilingual multi-step reasoning in the language models cannot be fully captured by human-like step-by-step sub-question answering, underscoring the need for future work to uncover the underlying mechanisms of model reasoning.
%On the other hand, this work demonstrates the benefits of explicit step-by-step prompting with structured sub-questions in multilingual multi-hop tasks, particularly for mitigating compositional failures. 
%Although the model is capable of reasoning without predefined structures, our findings show that such guidance enhances its robustness.
%These insights open promising directions for future work on incorporating reasoning decomposition into both training objectives and prompting strategies to improve multilingual multi-hop reasoning.

\section*{Limitations}

We acknowledge several limitations in our work.
First, this paper primarily analyzes how language models reason across languages.
The proposed \textsc{SUBQ} method is intended as a diagnostic tool for examining whether LLMs benefit from structured, step-by-step reasoning, rather than as a directly deployable method.
Its applicability is limited because our setting provides oracle decomposed sub-questions, whereas real-world settings require an additional question-decomposition process.
%Future work can investigate automatic decomposition methods and evaluate their effectiveness. 
Second, our translated dataset focuses on four high-resource languages and does not include low-resource languages. 
This design choice aims to mitigate the impact of training data scarcity on models' cross-lingual reasoning performance. 
Although these four high-resource languages still have less pre-training data compared to English, we believe they provide a sufficiently controlled setting to analyze cross-lingual reasoning without the extreme data imbalance in low-resource languages.

\section*{Acknowledgments}
This research was funded in part by the Netherlands Organization for Scientific Research (NWO)
under project number VI.C.192.080. We would like to thank Vlad Niculae, David Stap, and Sergey Troshin for their useful suggestions. We would also like to thank the reviewers for their feedback.

\bibliography{custom}
\newpage

\appendix

\section{Use of Large Language Models}
\label{app:llm-use}

We use large language models (LLMs) in three aspects: First, we use \texttt{GPT-4o-mini} to translate the dataset into the target languages. Second, we use \texttt{ChatGPT} for grammar-level polishing of the manuscript. Third, we use \texttt{Claude-OPUS-4.6} for error annotation. 
We do not use LLMs for research ideation, experimental design, or generating substantive scientific content.

\section{Translation Quality Evaluation}
\label{app:translation-quality}

We evaluate translation quality using both automatic and human evaluation. For automatic evaluation, we report reference-free \texttt{COMETKiwi}\footnote{\url{https://huggingface.co/Unbabel/wmt22-cometkiwi-da}} scores~\citep{rei-etal-2022-cometkiwi}. For human evaluation, we randomly sample $20\%$ of the translated dataset for each language. 
We invite PhD students at the University of Amsterdam to rate the translated dataset. 
They are native speakers of French, Chinese, Arabic, and Russian but also speak English fluently. 
They volunteered to participate in the translation rating task and therefore received no compensation.
They were informed of the purpose of the translation-rating task and were instructed to evaluate the dataset using the four-point scale described below.
Results are in Table \ref{tab:quality}. 

\begin{itemize}
    \item \textbf{3}: The translation preserves the full meaning of the English source and contains no grammatical errors.
    \item \textbf{2}: The translation preserves most of the meaning of the English source, with only minor grammatical errors.
    \item \textbf{1}: The translation preserves only part of the source meaning and may contain fluency or grammatical issues.
    \item \textbf{0}: The translation preserves little or none of the source meaning and is hard to understand.
\end{itemize}

\begin{table*}[!t]
    \centering
    \begin{tabular}{lcccc}
    \toprule
        \textbf{Metric} & \textbf{French (Fr)} & \textbf{Russian (Ru)} & \textbf{Chinese (Zh)} & \textbf{Arabic (Ar)} \\
        \midrule
        \texttt{COMETKiwi} & 86.14 & 83.42 & 82.89 & 80.43 \\
        Human & 2.60 & 2.40 & 2.50 & 2.60 \\
        \bottomrule
    \end{tabular}
    \caption{Translation quality of the multilingual two-hop QA dataset. We report reference-free \texttt{COMETKiwi} scores and average human ratings on a 0--3 scale. Both automatic and human evaluations indicate that the translations largely preserve the meaning of the English source questions and documents.}
    \label{tab:quality}
\end{table*}

\section{Linguistic Similarity Calculation}
\label{app:linguistic-similarity}

We compute linguistic similarity using subword vocabulary overlap on the NTREX multi-parallel corpus, \footnote{\url{https://github.com/MicrosoftTranslator/NTREX}} following \citet{Qi2025OnTC}. NTREX covers 128 languages and provides parallel text for estimating vocabulary overlap across languages. For two languages $l_1$ and $l_2$, we define the overlap score as:
\begin{equation}
    \text{Overlap}(l_1, l_2) = \frac{|V_1 \cap V_2|}{|V_1 \cup V_2|},
\end{equation}
where $V_1$ and $V_2$ denote the subword vocabularies induced from languages $l_1$ and $l_2$, respectively.

\section{Prompts}
\label{app:prompts}

\begin{figure*}[!t]
    \centering
    \begin{subfigure}[b]{\textwidth}
        \centering
        \includegraphics[width=\linewidth]{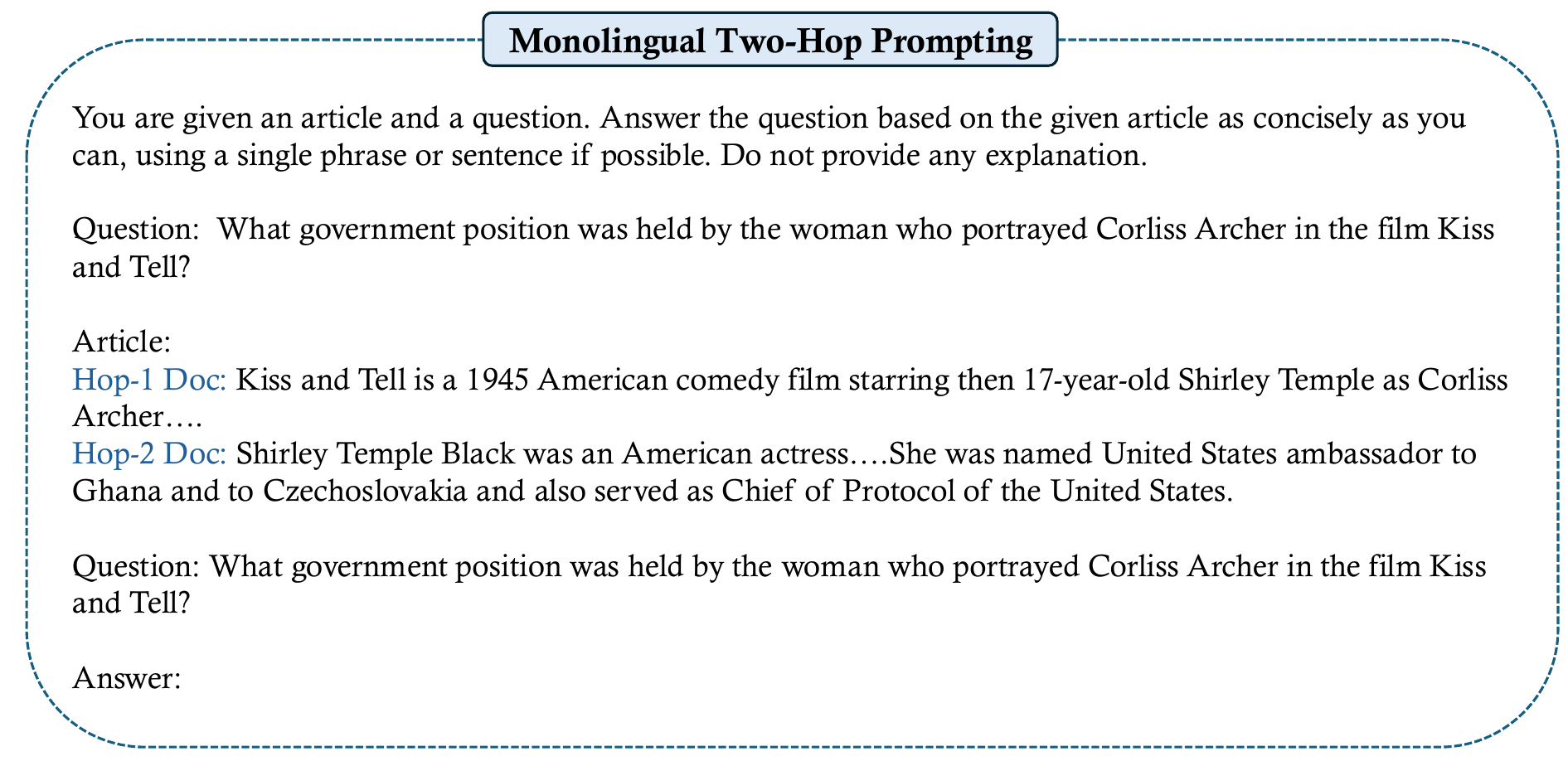}
        \caption{Monolingual setting: English query, English Hop-1 document, and English Hop-2 document.}
        \label{fig:prompt_en}
    \end{subfigure}
    \vspace{0.6em}
    \begin{subfigure}[b]{\textwidth}
        \centering
        \includegraphics[width=\linewidth]{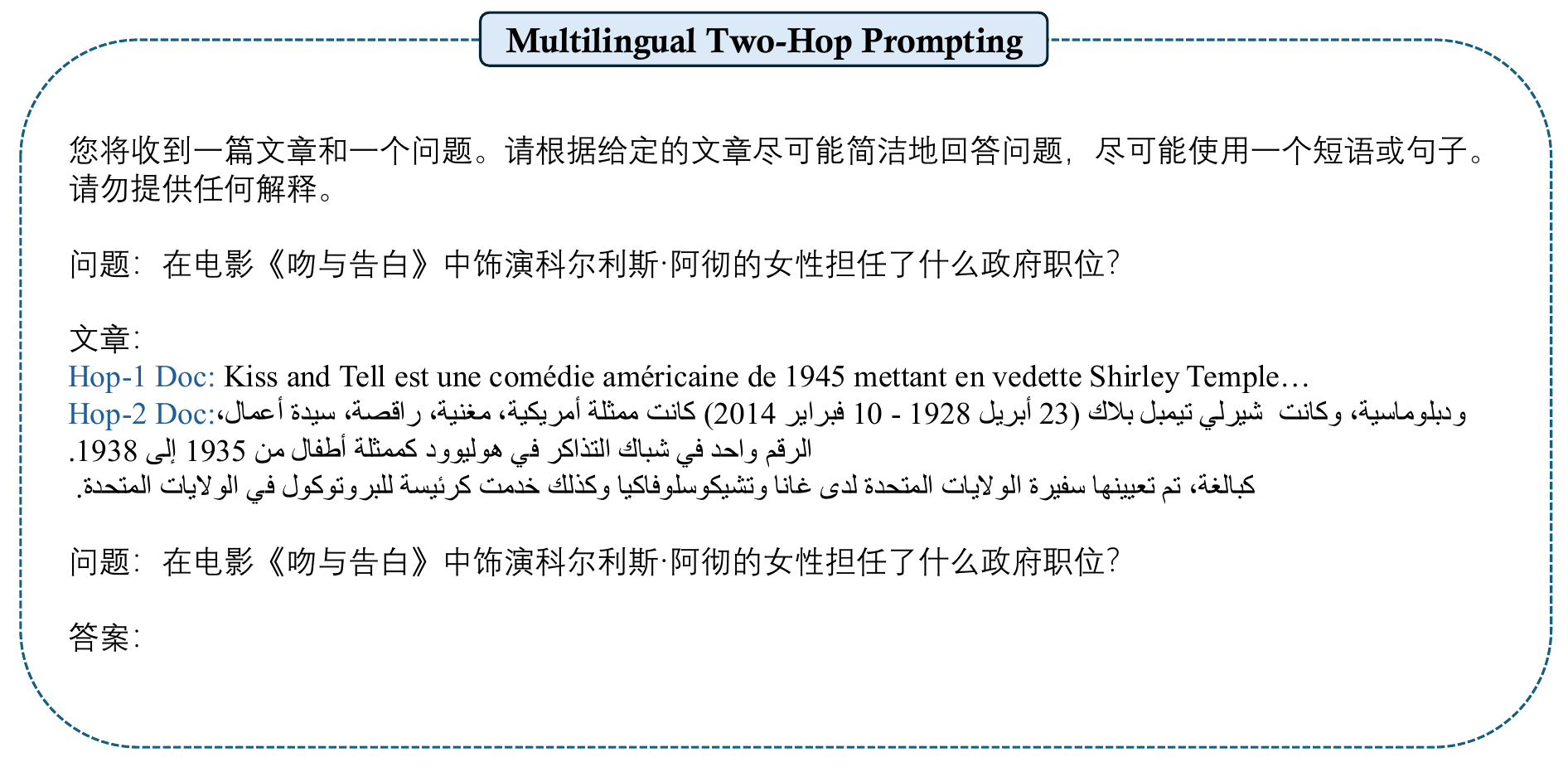}
        \caption{Cross-lingual setting: Chinese query, French Hop-1 document, and Arabic Hop-2 document.}
        \label{fig:prompt_zh}
    \end{subfigure}
    \caption{Default multilingual two-hop QA prompt templates. The labels \textcolor{blue}{Hop-1 Doc} and \textcolor{blue}{Hop-2 Doc} are shown only for illustration and are not included in the actual prompt.}
    \label{fig:default_prompt}
\end{figure*}

\begin{figure*}[!t]
    \centering
    \begin{subfigure}[b]{0.49\textwidth}
        \centering
        \includegraphics[width=\linewidth]{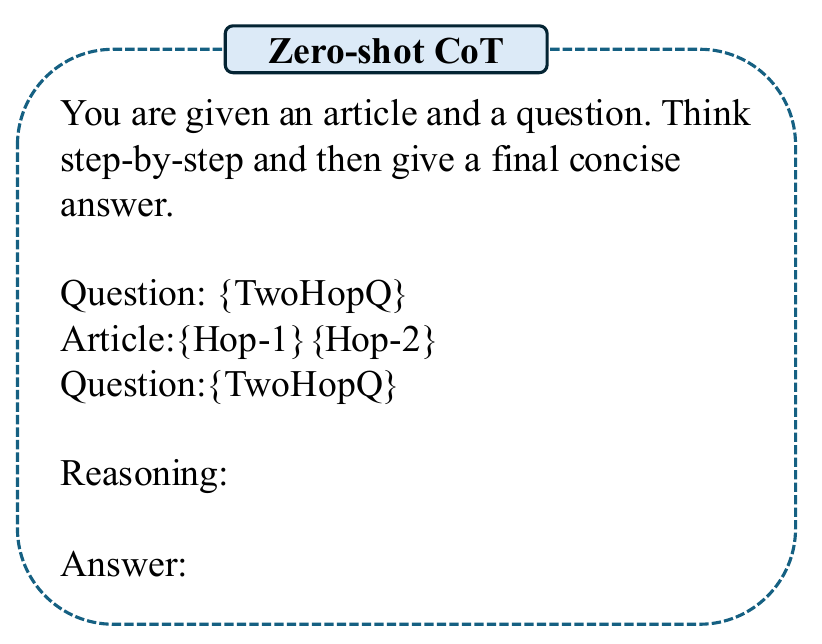}
        \caption{English two-hop question.}
    \end{subfigure}
    \hfill
    \begin{subfigure}[b]{0.49\textwidth}
        \centering
        \includegraphics[width=\linewidth]{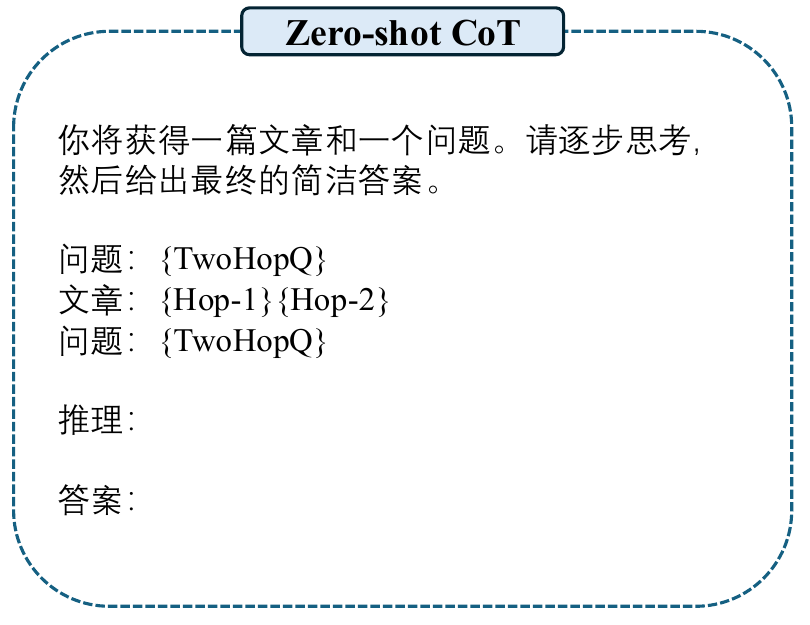}
        \caption{Chinese two-hop question.}
    \end{subfigure}
    \caption{Zero-shot CoT prompt templates. We append ``think step-by-step'' to encourage the model to generate an intermediate reasoning chain before answering.}
    \label{fig:zero-shot-cot}
\end{figure*}

Figure \ref{fig:default_prompt} and \ref{fig:zero-shot-cot} present prompt templates used for different experiments in this paper. 
In all settings, the instruction language is matched to the query language, with instructions translated by native speakers for each target language.

\section{Error Analysis}
  \label{sec:error-analysis}

\begin{itemize}\itemsep0pt\parsep1pt
  \item \textbf{Correct:} The prediction exactly matches the gold answer (i.e., Token-level F$1$ is 1).
  
\item \textbf{Partial Answer:} The prediction identifies the correct
  answer but is incomplete or overly specific. For example, given the gold answer ``Erick Sermon, Redman and Keith Murray,'' predicting only
  ``Erick Sermon''; similarly, predicting ``1950s'' for
  ``From the 1950s to the 1970s''.

  \item \textbf{Imprecise Translation:} The prediction identifies the
  correct underlying answer but contains translation errors in the target language. 

  \item \textbf{Factual Errors:} The prediction is in the correct language
  but gives factually incorrect information. 
  
\item \textbf{Wrong Language:} The prediction is in a language different from the query language, regardless of whether the underlying answer is factually correct. 

\end{itemize}

\section{Additional Model Results}
\label{app:additional-models}

Tables~\ref{tab:combined-2hop-qwen}, \ref{tab:combined-2hop-aya}, and \ref{tab:combined-2hop-llama} report additional results for Qwen-3-8B, Aya-Expanse-8B and Llama-3.1-8B-Instruct.

\section{Full Multilingual Two-hop QA Results}
\label{app:full-multilingual-results}

\begin{figure*}[!t]
    \centering
    \begin{subfigure}[b]{0.45\textwidth}
        \centering
        \includegraphics[width=\linewidth]{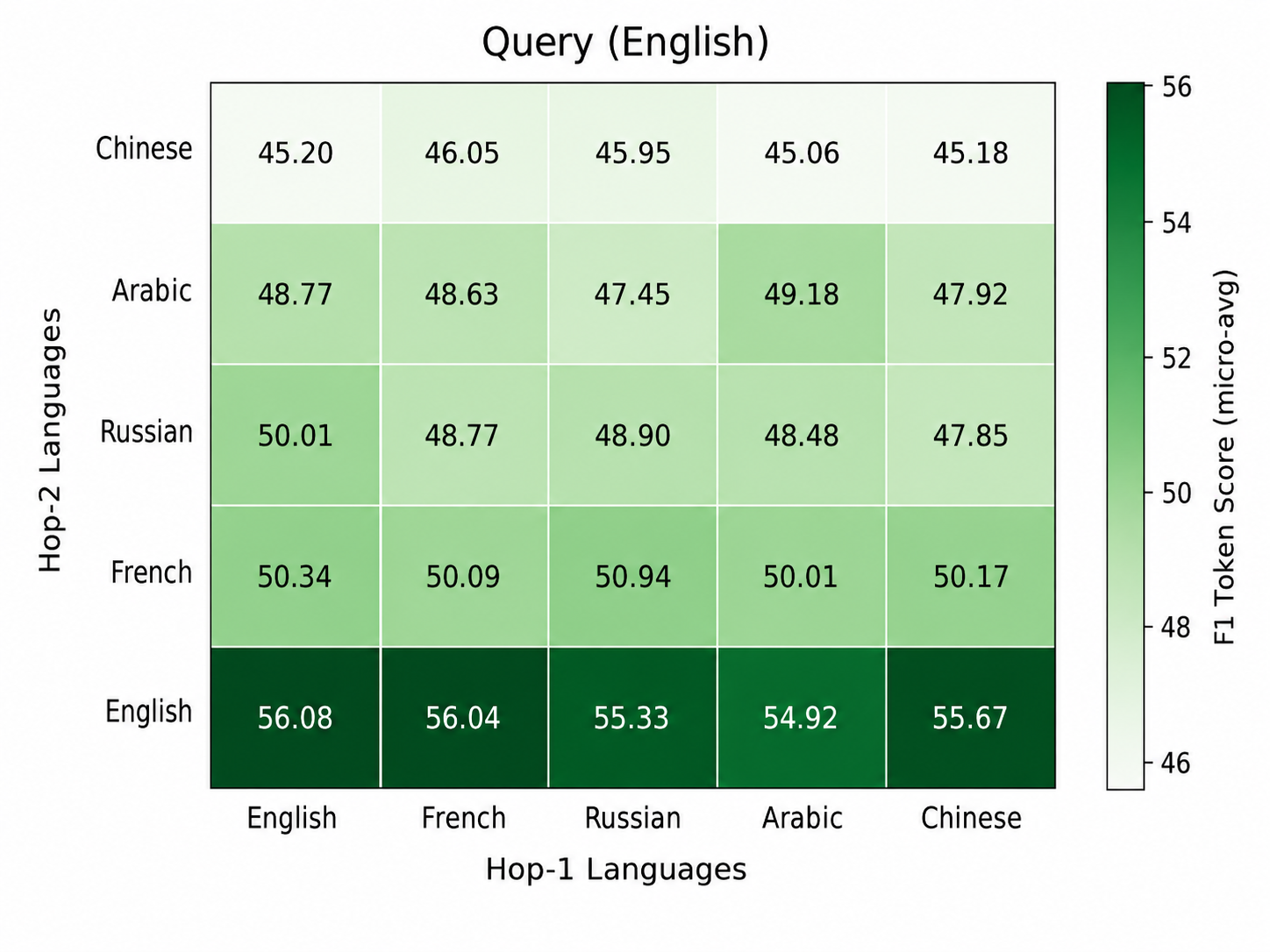}
        \caption{English}
    \end{subfigure}
    \hfill
    \begin{subfigure}[b]{0.45\textwidth}
        \centering
        \includegraphics[width=\linewidth]{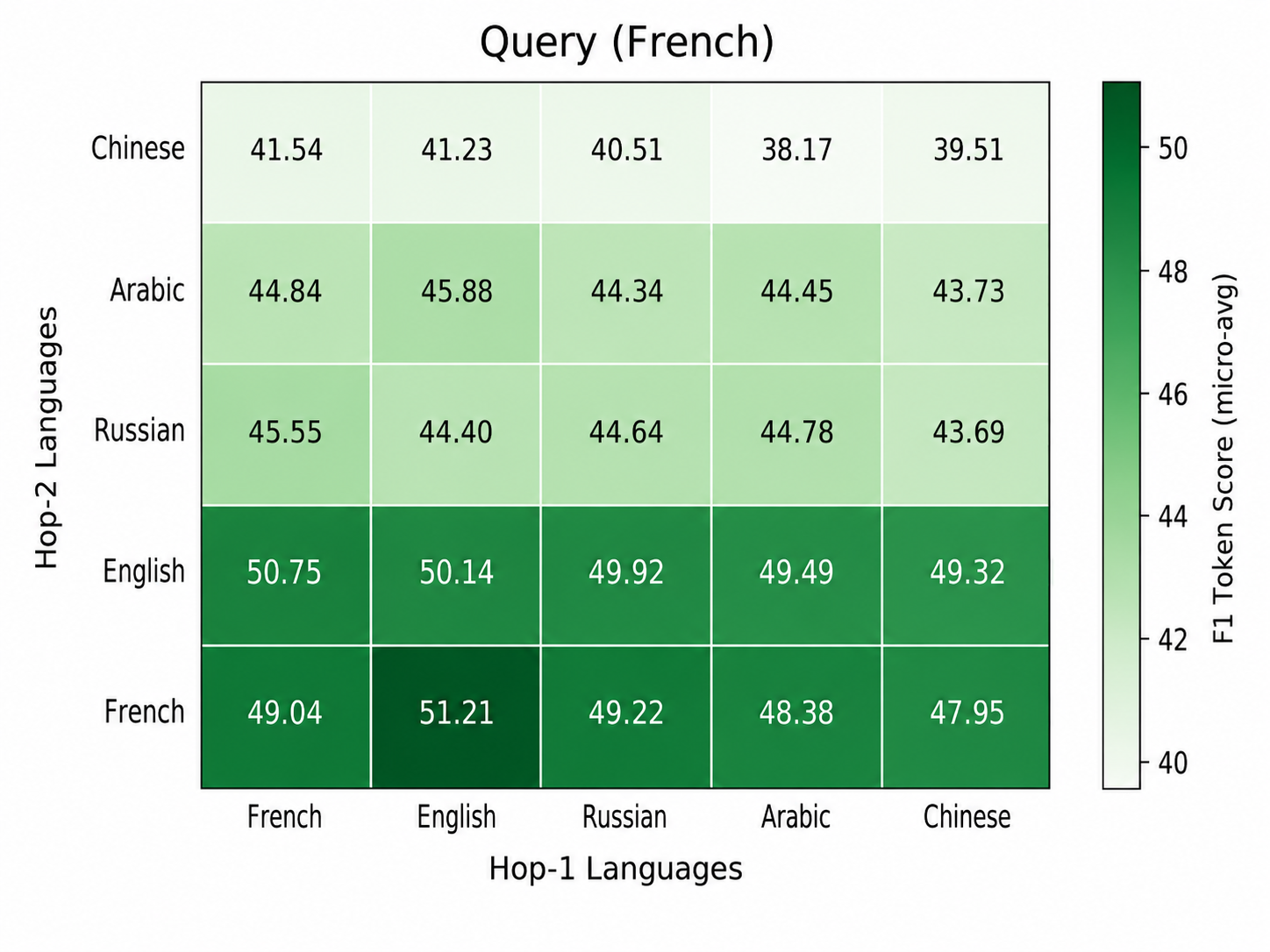}
        \caption{French}
    \end{subfigure}

    \vspace{0.7em}

    \begin{subfigure}[b]{0.45\textwidth}
        \centering
        \includegraphics[width=\linewidth]{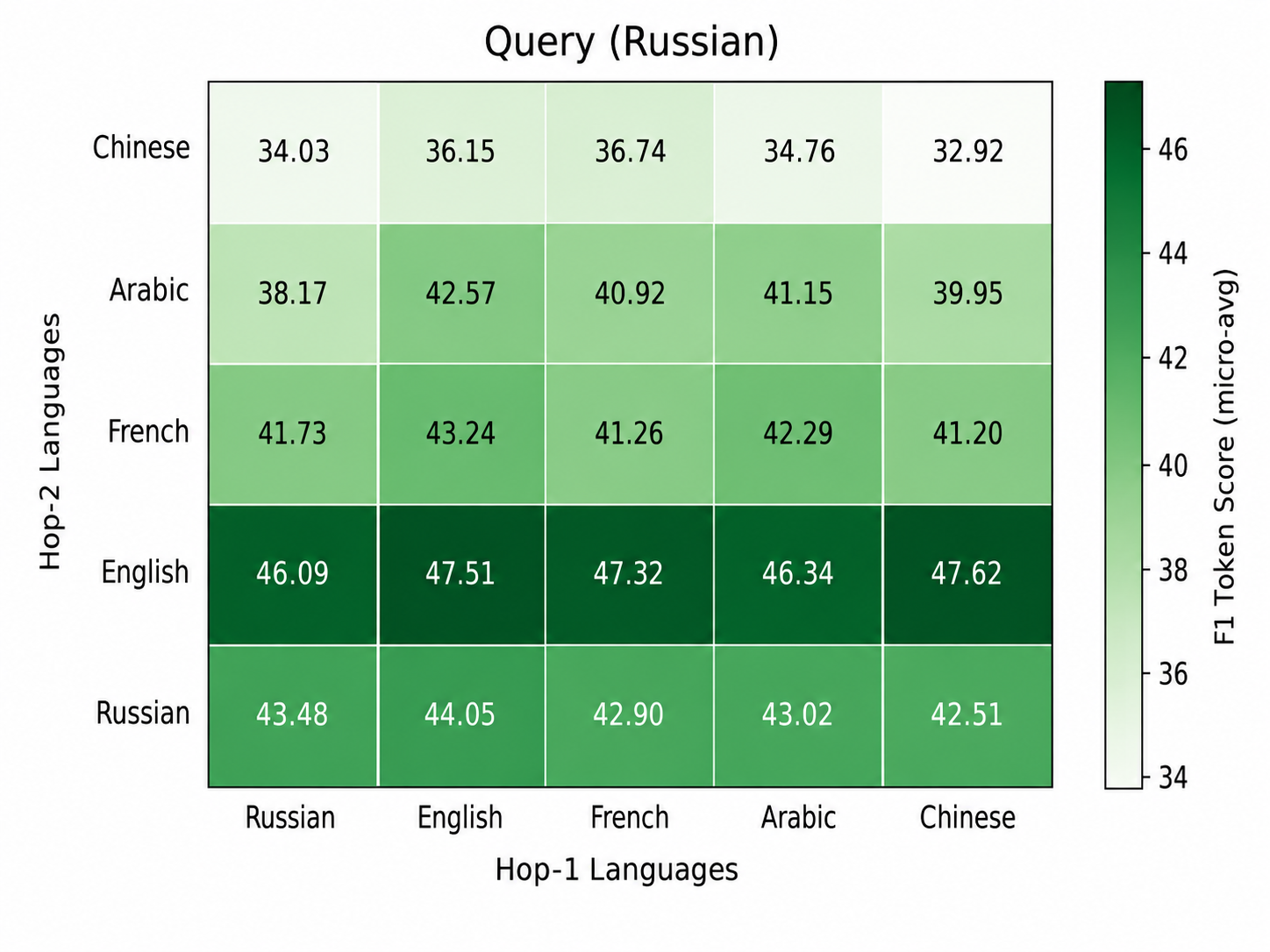}
        \caption{Russian}
    \end{subfigure}
    \hfill
    \begin{subfigure}[b]{0.45\textwidth}
        \centering
        \includegraphics[width=\linewidth]{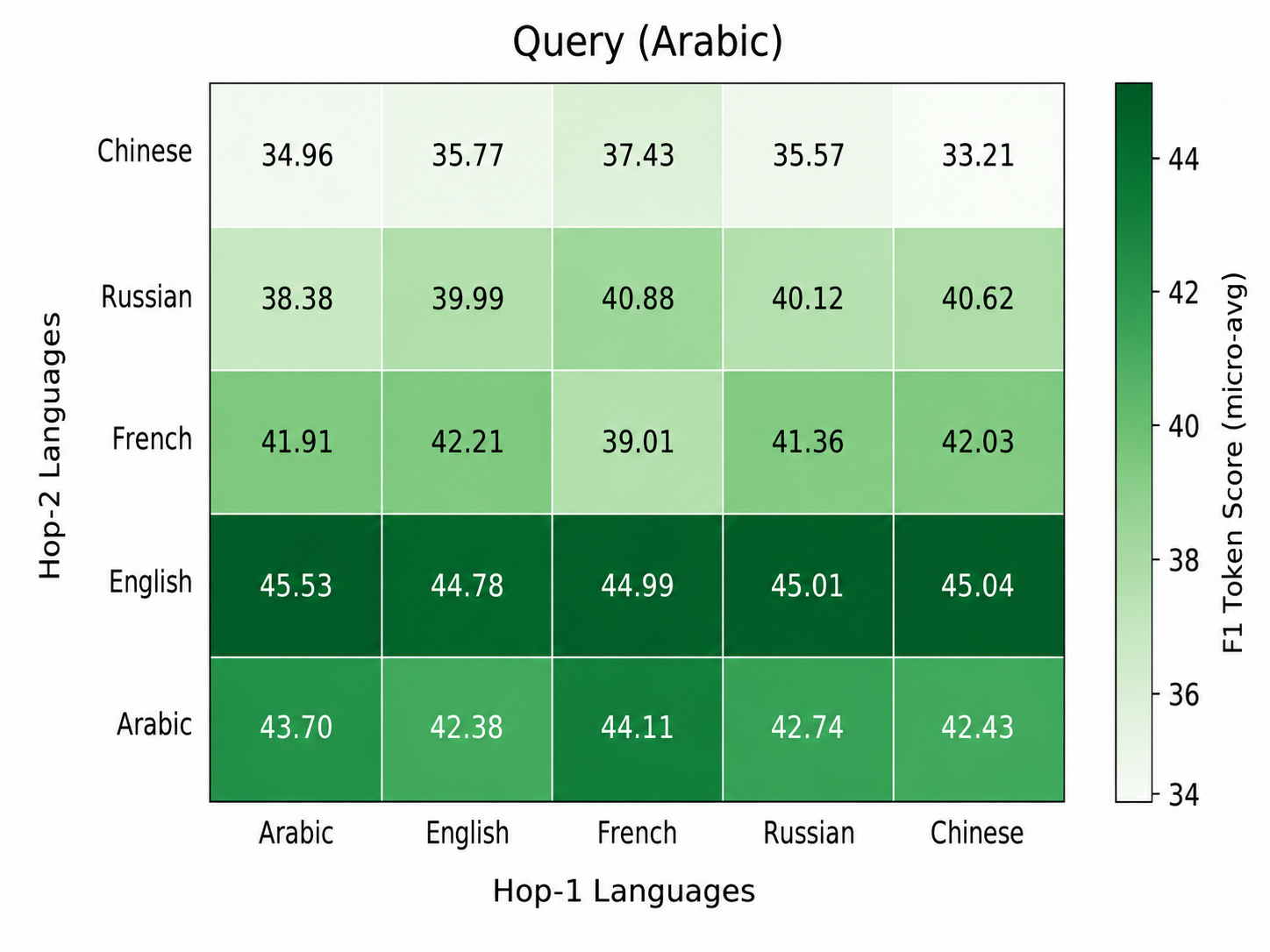}
        \caption{Arabic}
    \end{subfigure}

    \vspace{0.7em}

    \begin{subfigure}[b]{0.45\textwidth}
        \centering
        \includegraphics[width=\linewidth]{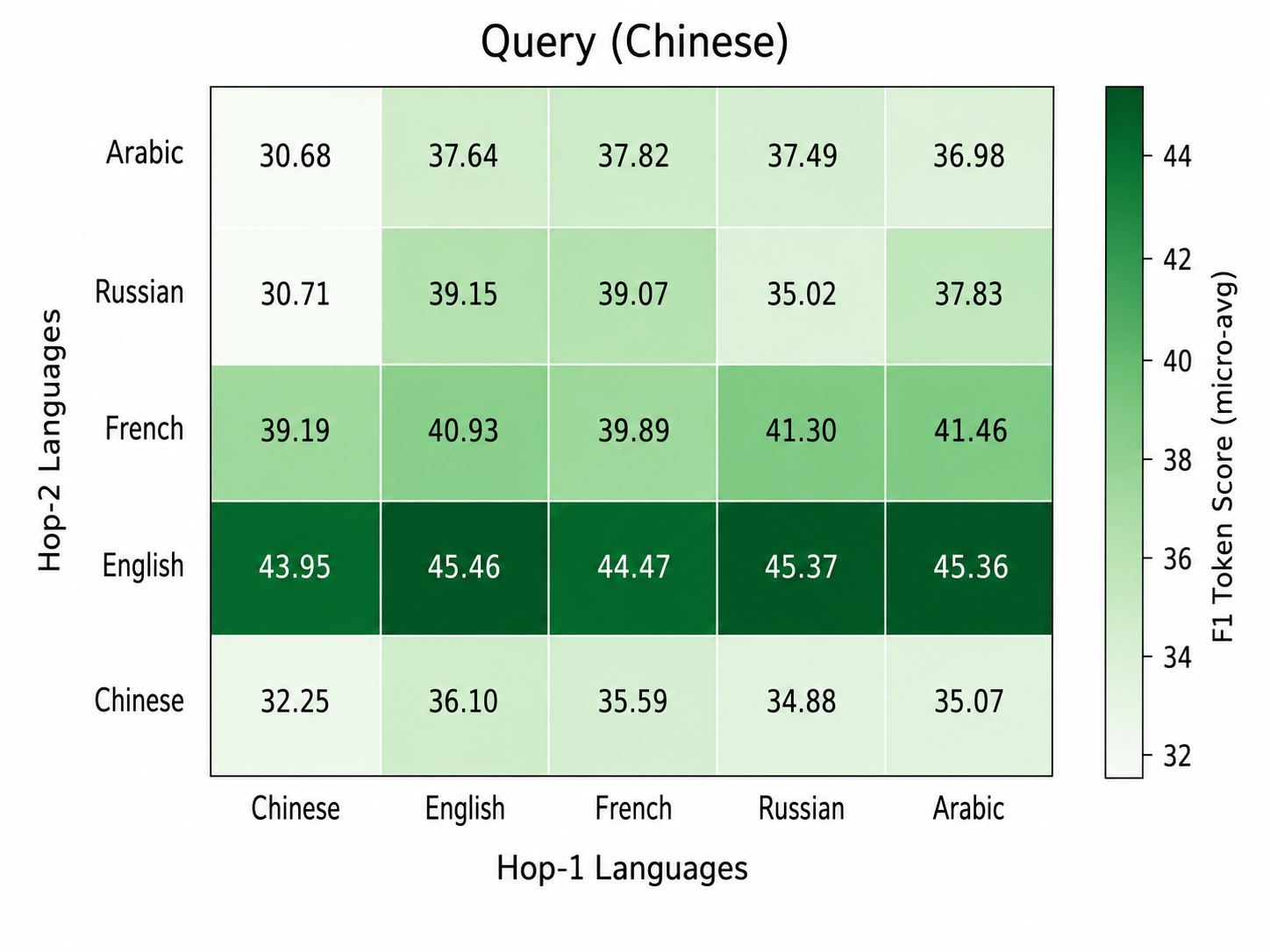}
        \caption{Chinese}
    \end{subfigure}

    \caption{Full multilingual two-hop QA performance for Gemma-3-27B-Instruct across query languages. Each panel fixes the query language and varies the languages of the Hop-1 and Hop-2 evidence documents.}
    \label{fig:27-B-langs}
\end{figure*}

Figure~\ref{fig:27-B-langs} presents the full multilingual two-hop QA results for Gemma-3-27B-Instruct across five query languages under different multilingual controlled settings.

\begin{table*}[!t]
  \centering
  \resizebox{\linewidth}{!}{%
      \begin{tabular}{lllcccccc}
        \toprule
        \multirow{2}{*}{\textbf{Setting}} & \multicolumn{2}{c}{\textbf{Document Languages}} & \multicolumn{5}{c}{\textbf{Query Languages ($L_q$)}} & \multirow{2}{*}{\textbf{Avg.}} \\
        \cmidrule(lr){2-3} \cmidrule(lr){4-8}
        & \textbf{Hop-1} & \textbf{Hop-2} & \textbf{En} & \textbf{Fr} & \textbf{Ru} & \textbf{Ar} & \textbf{Zh} & \\
        \midrule
        \textbf{Monolingual} & 
        %$Q$ & $Q$ & 
        $=L_q$ & $=L_q$ & 
        \colorbox{e8}{52.59} & \colorbox{e12}{45.64} & \colorbox{e15}{38.62} & \colorbox{e16}{37.73} & \colorbox{e20}{27.47} & 40.41 \\
        \hdashline
        \multirow{3}{*}{\textbf{Cross-lingual}} & 
        %$\overline{Q}$ & $Q$ & 
        $\ne L_q$ & $=L_q$ & 
        \colorbox{e8}{52.99} & \colorbox{e12}{44.56} & \colorbox{e15}{38.63} & \colorbox{e16}{37.23} & \colorbox{e20}{25.84} & 39.85 \\
        %& $Q$ & $\overline{Q}$ & 
        & $=L_q$ & $\ne L_q$ & 
        \colorbox{e14}{41.80} & \colorbox{e16}{36.30} & \colorbox{e20}{29.19} & \colorbox{e20}{27.25} & \colorbox{e20}{23.89} & 31.69 \\
        & 
        %$\overline{Q}$ & $\overline{Q}$ & 
        $\ne L_q$ & $\ne L_q$ &         
        \colorbox{e15}{39.78} & \colorbox{e17}{34.79} & \colorbox{e20}{28.40} & \colorbox{e20}{26.89} & \colorbox{e20}{24.12} & 30.80 \\
        \bottomrule
      \end{tabular}}
    \caption{Multilingual two-hop QA performance for Qwen-3-8B. We report token-level F1. $=L_q$ and $\ne L_q$ indicate whether the document language matches or differs from the query language. For non-matching documents, we report the average across the four non-query languages.}
  \label{tab:combined-2hop-qwen}
\end{table*}

\begin{table*}[!t]
  \centering
  \resizebox{\linewidth}{!}{%
      \begin{tabular}{lllcccccc}
        \toprule
        \multirow{2}{*}{\textbf{Setting}} & \multicolumn{2}{c}{\textbf{Document Languages}} & \multicolumn{5}{c}{\textbf{Query Languages ($L_q$)}} & \multirow{2}{*}{\textbf{Avg.}} \\
        \cmidrule(lr){2-3} \cmidrule(lr){4-8}
        & \textbf{Hop-1} & \textbf{Hop-2} & \textbf{En} & \textbf{Fr} & \textbf{Ru} & \textbf{Ar} & \textbf{Zh} & \\
        \midrule
        \textbf{Monolingual} & 
        %$Q$ & $Q$ & 
        $=L_q$ & $=L_q$ & 
        \colorbox{e8}{60.06} & \colorbox{e13}{43.30} & \colorbox{e18}{34.91} & \colorbox{e14}{42.08} & \colorbox{e17}{36.07} & 43.29 \\
        \hdashline
        \multirow{3}{*}{\textbf{Cross-lingual}} & 
        %$\overline{Q}$ & $Q$ & 
        $\ne L_q$ & $=L_q$ & 
        \colorbox{e8}{58.08} & \colorbox{e13}{43.41} & \colorbox{e18}{34.62} & \colorbox{e14}{41.56} & \colorbox{e18}{34.71} & 42.47 \\
        %& $Q$ & $\overline{Q}$ & 
        & $=L_q$ & $\ne L_q$ & 
        \colorbox{e20}{25.27} & \colorbox{e18}{33.12} & \colorbox{e19}{31.86} & \colorbox{e17}{36.35} & \colorbox{e20}{22.71} & 29.86 \\
        & 
        %$\overline{Q}$ & $\overline{Q}$ & 
        $\ne L_q$ & $\ne L_q$ &         
        \colorbox{e20}{19.94} & \colorbox{e19}{31.88} & \colorbox{e19}{31.41} & \colorbox{e18}{34.89} & \colorbox{e20}{18.82} & 27.39 \\
        \bottomrule
      \end{tabular}}
    \caption{Multilingual two-hop QA performance for Aya-Expanse-8B. We report token-level F1. $=L_q$ and $\ne L_q$ indicate whether the document language matches or differs from the query language. For non-matching documents, we report the average across the four non-query languages.}
  \label{tab:combined-2hop-aya}
\end{table*}

\begin{table*}[!t]
  \centering
  \resizebox{\linewidth}{!}{%
      \begin{tabular}{lllcccccc}
        \toprule
        \multirow{2}{*}{\textbf{Setting}} & \multicolumn{2}{c}{\textbf{Document Languages}} & \multicolumn{5}{c}{\textbf{Query Languages ($L_q$)}} & \multirow{2}{*}{\textbf{Avg.}} \\
        \cmidrule(lr){2-3} \cmidrule(lr){4-8}
        & \textbf{Hop-1} & \textbf{Hop-2} & \textbf{En} & \textbf{Fr} & \textbf{Ru} & \textbf{Ar} & \textbf{Zh} & \\
        \midrule
        \textbf{Monolingual} & 
        %$Q$ & $Q$ & 
        $=L_q$ & $=L_q$ & 
        \colorbox{e11}{46.64} & \colorbox{e15}{39.07} & \colorbox{e20}{29.24} & \colorbox{e20}{28.30} & \colorbox{e20}{23.24} & 33.30 \\
        \hdashline
        \multirow{3}{*}{\textbf{Cross-lingual}} & 
        %$\overline{Q}$ & $Q$ 
        $\ne L_q$ & $=L_q$ 
        & \colorbox{e13}{43.91} & \colorbox{e16}{36.21} & \colorbox{e20}{28.60} & \colorbox{e20}{24.62} & \colorbox{e20}{21.90} & 31.05 \\
        %& $Q$ & $\overline{Q}$ 
        & $=L_q$ & $\ne L_q$ 
        & \colorbox{e17}{35.49} & \colorbox{e19}{30.55} & \colorbox{e20}{22.10} & \colorbox{e20}{18.64} & \colorbox{e20}{14.67} & 24.29 \\
        %& $\overline{Q}$ & $\overline{Q}$ &
        & $\ne L_q$ & $\ne L_q$ &
        \colorbox{e19}{30.10} & \colorbox{e20}{28.57} & \colorbox{e20}{22.27} & \colorbox{e20}{18.86} & \colorbox{e20}{16.47} & 23.25 \\
        \bottomrule
      \end{tabular}}
    \caption{Multilingual two-hop QA performance for Llama-3.1-8B-Instruct. We report token-level F1. $=L_q$ and $\ne L_q$ indicate whether the document language matches or differs from the query language. For non-matching documents, we report the average across the four non-query languages.}
  \label{tab:combined-2hop-llama}
\end{table*}

\end{document}